\pdfoutput=1
\documentclass[10pt,twocolumn,letterpaper]{article}

\usepackage{3dv}
\usepackage{times}
\usepackage{epsfig}
\usepackage{graphicx}
\usepackage{amsmath}
\usepackage{amssymb}

\usepackage{booktabs}
\usepackage[separate-uncertainty=true]{siunitx}
\usepackage[inline]{enumitem}
\usepackage{mathtools}
\usepackage{multirow}
\usepackage{multicol}
\usepackage{changepage} 
\graphicspath{{gfx/}}
\usepackage{soul}
\usepackage{tabularx}
\usepackage{adjustbox}
\usepackage{float}

\usepackage[pagebackref=true,breaklinks=true,letterpaper=true,colorlinks,bookmarks=false]{hyperref}

\threedvfinalcopy 

\def\threedvPaperID{253} 

\begin{document}

\title{Self-Supervised Monocular Scene Decomposition and Depth Estimation}

\author{Sadra Safadoust \qquad Fatma G\"uney\\
KUIS AI Center, Ko\c{c} University\\
{\tt\small ssafadoust20@ku.edu.tr} \qquad {\tt\small fguney@ku.edu.tr}
}

\maketitle

\newcommand{\Perp}{\perp\!\!\! \perp}
\newcommand{\bK}{\mathbf{K}}
\newcommand{\bX}{\mathbf{X}}
\newcommand{\bY}{\mathbf{Y}}
\newcommand{\bk}{\mathbf{k}}
\newcommand{\bx}{\mathbf{x}}
\newcommand{\by}{\mathbf{y}}
\newcommand{\bhy}{\hat{\mathbf{y}}}
\newcommand{\bty}{\tilde{\mathbf{y}}}
\newcommand{\bG}{\mathbf{G}}
\newcommand{\bI}{\mathbf{I}}
\newcommand{\bg}{\mathbf{g}}
\newcommand{\bS}{\mathbf{S}}
\newcommand{\bs}{\mathbf{s}}
\newcommand{\bM}{\mathbf{M}}
\newcommand{\bw}{\mathbf{w}}
\newcommand{\eye}{\mathbf{I}}
\newcommand{\bU}{\mathbf{U}}
\newcommand{\bV}{\mathbf{V}}
\newcommand{\bW}{\mathbf{W}}
\newcommand{\bn}{\mathbf{n}}
\newcommand{\bv}{\mathbf{v}}
\newcommand{\bwv}{\mathbf{wv}}
\newcommand{\bq}{\mathbf{q}}
\newcommand{\bR}{\mathbf{R}}
\newcommand{\bi}{\mathbf{i}}
\newcommand{\bj}{\mathbf{j}}
\newcommand{\bp}{\mathbf{p}}
\newcommand{\bt}{\mathbf{t}}
\newcommand{\bJ}{\mathbf{J}}
\newcommand{\bu}{\mathbf{u}}
\newcommand{\bB}{\mathbf{B}}
\newcommand{\bD}{\mathbf{D}}
\newcommand{\bz}{\mathbf{z}}
\newcommand{\bP}{\mathbf{P}}
\newcommand{\bC}{\mathbf{C}}
\newcommand{\bA}{\mathbf{A}}
\newcommand{\bZ}{\mathbf{Z}}
\newcommand{\bff}{\mathbf{f}}
\newcommand{\bF}{\mathbf{F}}
\newcommand{\bo}{\mathbf{o}}
\newcommand{\bO}{\mathbf{O}}
\newcommand{\bc}{\mathbf{c}}
\newcommand{\bm}{\mathbf{m}}
\newcommand{\bT}{\mathbf{T}}
\newcommand{\bQ}{\mathbf{Q}}
\newcommand{\bL}{\mathbf{L}}
\newcommand{\bl}{\mathbf{l}}
\newcommand{\ba}{\mathbf{a}}
\newcommand{\bE}{\mathbf{E}}
\newcommand{\bH}{\mathbf{H}}
\newcommand{\bd}{\mathbf{d}}
\newcommand{\br}{\mathbf{r}}
\newcommand{\be}{\mathbf{e}}
\newcommand{\bb}{\mathbf{b}}
\newcommand{\bh}{\mathbf{h}}
\newcommand{\bhh}{\hat{\mathbf{h}}}
\newcommand{\btheta}{\boldsymbol{\theta}}
\newcommand{\bTheta}{\boldsymbol{\Theta}}
\newcommand{\bpi}{\boldsymbol{\pi}}
\newcommand{\bphi}{\boldsymbol{\phi}}
\newcommand{\bPhi}{\boldsymbol{\Phi}}
\newcommand{\bmu}{\boldsymbol{\mu}}
\newcommand{\bSigma}{\boldsymbol{\Sigma}}
\newcommand{\bGamma}{\boldsymbol{\Gamma}}
\newcommand{\bbeta}{\boldsymbol{\beta}}
\newcommand{\bomega}{\boldsymbol{\omega}}
\newcommand{\blambda}{\boldsymbol{\lambda}}
\newcommand{\bLambda}{\boldsymbol{\Lambda}}
\newcommand{\bkappa}{\boldsymbol{\kappa}}
\newcommand{\btau}{\boldsymbol{\tau}}
\newcommand{\balpha}{\boldsymbol{\alpha}}
\newcommand{\nR}{\mathbb{R}}
\newcommand{\nN}{\mathbb{N}}
\newcommand{\nL}{\mathbb{L}}
\newcommand{\cN}{\mathcal{N}}
\newcommand{\cM}{\mathcal{M}}
\newcommand{\cR}{\mathcal{R}}
\newcommand{\cB}{\mathcal{B}}
\newcommand{\cL}{\mathcal{L}}
\newcommand{\cH}{\mathcal{H}}
\newcommand{\cS}{\mathcal{S}}
\newcommand{\cT}{\mathcal{T}}
\newcommand{\cO}{\mathcal{O}}
\newcommand{\cC}{\mathcal{C}}
\newcommand{\cP}{\mathcal{P}}
\newcommand{\cE}{\mathcal{E}}
\newcommand{\cI}{\mathcal{I}}
\newcommand{\cF}{\mathcal{F}}
\newcommand{\cK}{\mathcal{K}}
\newcommand{\cY}{\mathcal{Y}}
\newcommand{\cX}{\mathcal{X}}
\def\bgamma{\boldsymbol\gamma}

\newcommand{\specialcell}[2][c]{%
  \begin{tabular}[#1]{@{}c@{}}#2\end{tabular}}

\newcommand{\figref}[1]{\Fig~\ref{#1}}
\newcommand{\secref}[1]{Section~\ref{#1}}
\newcommand{\algref}[1]{Algorithm~\ref{#1}}
\newcommand{\eqnref}[1]{Eq.~\eqref{#1}}
\newcommand{\tabref}[1]{Table~\ref{#1}}

\newcommand{\rulesep}{\unskip\ \vrule\ }



\newcommand{\KLD}[2]{D_{\mathrm{KL}} \left( \left. \left. #1 \right|\right| #2 \right) }

\renewcommand{\b}{\ensuremath{\mathbf}}

\def\mc{\mathcal}
\def\mb{\mathbf}

\newcommand{\T}{^{\raisemath{-1pt}{\mathsf{T}}}}

\makeatletter
\DeclareRobustCommand\onedot{\futurelet\@let@token\@onedot}
\def\@onedot{\ifx\@let@token.\else.\null\fi\xspace}
\def\eg{e.g\onedot} \def\Eg{E.g\onedot}
\def\ie{i.e\onedot} \def\Ie{I.e\onedot}
\def\cf{cf\onedot} \def\Cf{Cf\onedot}
\def\etc{etc\onedot} \def\vs{vs\onedot}
\def\wrt{wrt\onedot}
\def\dof{d.o.f\onedot}
\def\etal{et~al\onedot} \def\iid{i.i.d\onedot}
\def\Fig{Fig\onedot} \def\Eqn{Eqn\onedot} \def\Sec{Sec\onedot} \def\Alg{Alg\onedot}
\makeatother

\newcommand{\xdownarrow}[1]{%
  {\left\downarrow\vbox to #1{}\right.\kern-\nulldelimiterspace}
}

\newcommand{\xuparrow}[1]{%
  {\left\uparrow\vbox to #1{}\right.\kern-\nulldelimiterspace}
}

\renewcommand\UrlFont{\color{blue}\rmfamily}

\newcommand*\rot{\rotatebox{90}}
\newcommand{\boldparagraph}[1]{\vspace{0.2cm}\noindent{\bf #1:} }

\newcommand{\sadra}[1]{ \noindent {\color{blue} {\bf Sadra:} {#1}} }
\newcommand{\ftm}[1]{ \noindent {\color{magenta} {\bf Fatma:} {#1}} }

\newcommand{\supptitle}[1]{
   \newpage
   \null
  \vskip .375in
   \begin{center}
      {\Large \bf #1 \par}
      \vspace*{24pt}
      {
      \large
      \lineskip .5em
      \begin{tabular}[t]{c}
         Sadra Safadoust \qquad Fatma G\"uney\\
KUIS AI Center, Ko\c{c} University\\
{\tt\small ssafadoust20@ku.edu.tr} \qquad {\tt\small fguney@ku.edu.tr}\\
         \vspace*{1pt}\\
      \end{tabular}
      \par
      }
   \end{center}
   }
   
\newcommand{\titlesupp}[1]{
  \begin{table*}[Ht!]
   \vskip .375in
        
      {\Large \bf #1 \par}
        
      \vspace*{24pt}
      {
      \large
      \lineskip .5em
      \begin{tabular}{c}
         Sadra Safadoust \qquad Fatma G\"uney\\
KUIS AI Center, Ko\c{c} University\\
{\tt\small ssafadoust20@ku.edu.tr} \qquad {\tt\small fguney@ku.edu.tr}\\
         \vspace*{1pt}\\
      \end{tabular}
      
      \par
      }
      \vskip .5em
      \vspace*{12pt}
  \end{table*}
   }

\begin{abstract}
Self-supervised monocular depth estimation approaches either ignore independently moving objects in the scene or need a separate segmentation step to identify them.
We propose \textbf{MonoDepthSeg} to jointly estimate depth and segment moving objects from monocular video without using any ground-truth labels. 
We decompose the scene into a fixed number of components where each component corresponds to a region on the image with its own transformation matrix representing its motion. 
We estimate both the mask and the motion of each component efficiently with a shared encoder.
We evaluate our method on three driving datasets and show that our model clearly improves depth estimation while decomposing the scene into separately moving components.
\end{abstract}

\section{Introduction}
Humans are remarkably good at decomposing a scene into semantically meaningful components and inferring the 3D structure of the scene. This ability is crucial for self-driving vehicles while navigating complex environments to avoid collision. In computer vision, we often address the problems of inferring 3D structure and semantics of the scene separately. Years of research in geometric computer vision has perfected matching pixels and using this as a cue for self-supervision recently in deep learning. Yet, most of these methods fail to attribute a semantic meaning to discovered geometry with some notable exceptions \cite{Hoiem2005ICCV, Saxena2007IJCAI, Ladicky2014CVPR, Hane2014CVPR, Savinov2016CVPR}. Similarly, semantic segmentation of the scene often solely relies on RGB images while complicated occlusion relationships between objects can easily be untangled with the help of 3D information.

\begin{figure}[t]
    \label{fig:illustration}
    \def\imgw{0.49\linewidth}
    \def\imgh{0.15\linewidth}
    \def\hspacing{0.06cm}
    \def\vspacing{0.04cm}
    \begin{minipage}{\linewidth}
        \centering
        \includegraphics[width=\imgw,height=\imgh]{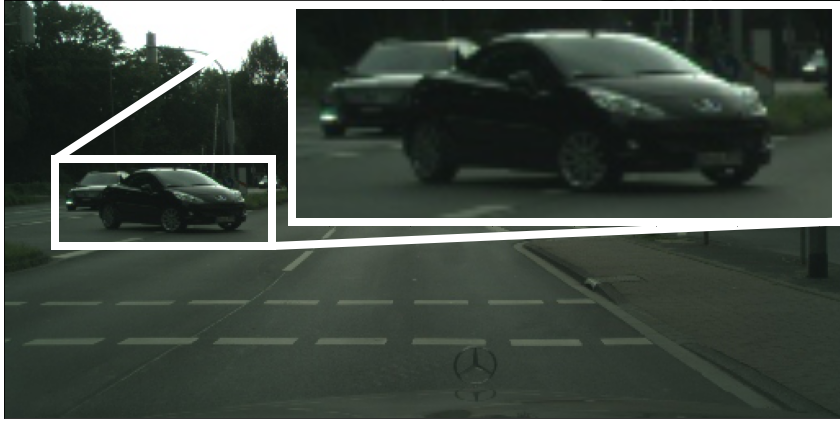}\hspace{\hspacing}%
        \includegraphics[width=\imgw,height=\imgh]{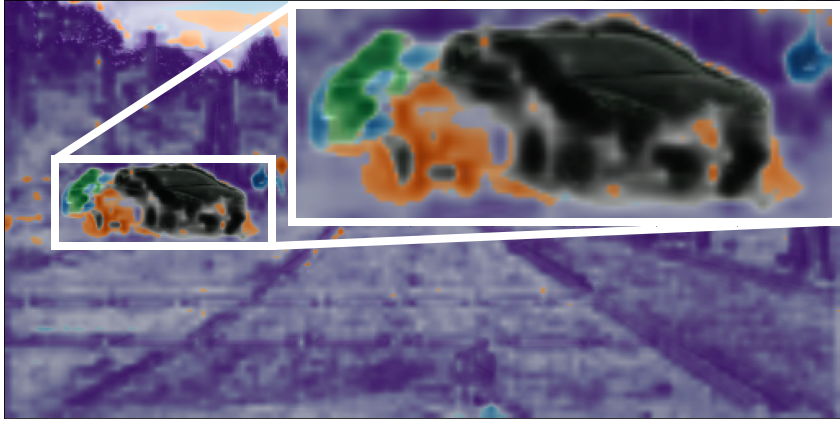}%
        \vspace{\vspacing}
        \includegraphics[width=\imgw,height=\imgh]{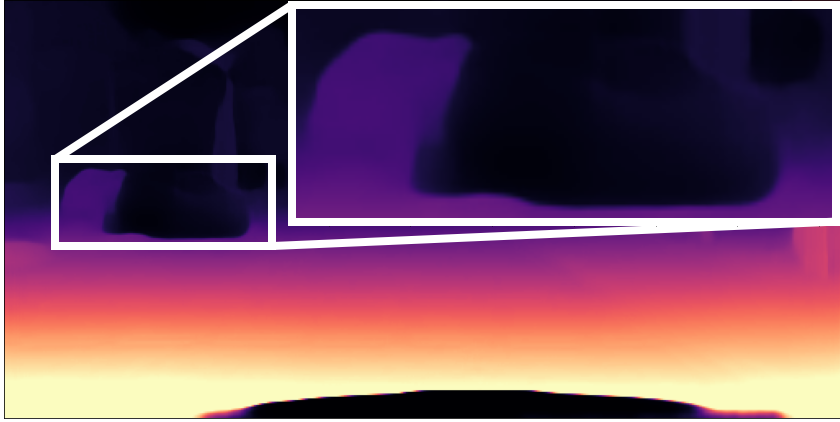}\hspace{\hspacing}%
        \includegraphics[width=\imgw,height=\imgh]{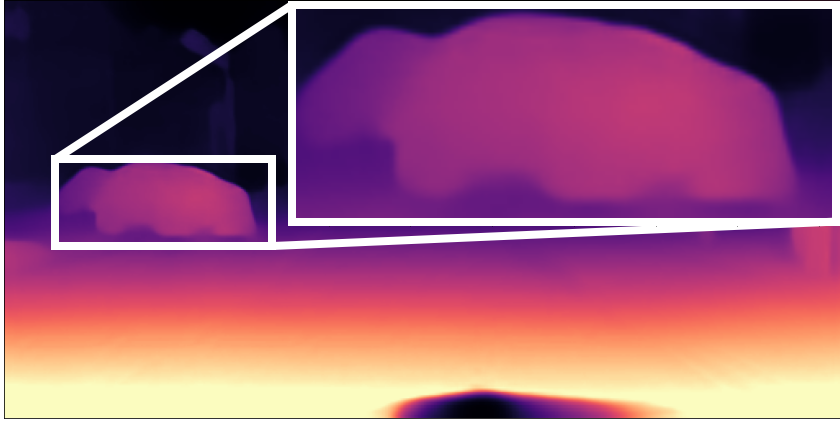}%
    \end{minipage}
    \vspace{0.1cm}
    \caption{\textbf{Joint Scene Decomposition and Depth Estimation.} Monocular depth estimation methods assume a static scene by relying on the ego-motion to explain the scene and fail in foreground regions with independently moving objects
    (\textbf{bottom-left}, \cite{Godard2019ICCV}). By decomposing the scene into a set of components, we estimate a separate rigid transformation for each component, representing its motion. This improves the results in regions with moving objects (\textbf{bottom-right}) while simultaneously recovering a decomposition of the scene, mostly corresponding to moving regions (\textbf{top-right}).}
\end{figure}

There has been a great progress in learning of scene structure from a moving camera without using any labels. Self-supervised models can be trained to accurately estimate depth and the ego-motion of the vehicle by matching regularities in the scene across time.
By following the progress in self-supervised monocular depth and ego-motion estimation \cite{Zhou2017CVPR, Godard2019ICCV}, we use view synthesis as an objective. In view synthesis, two models are trained to predict both the scene structure from a reference view and the relative ego-motion from the source view to the reference view. Then, the reference view can be synthesized by sampling pixels from the source view according to these estimations.
While monocular methods need to associate semantics of the scene with structure to estimate depth, semantics are not typically represented explicitly in the model.

The lack of semantics become apparent when there are independently moving objects whose motion cannot be explained by the ego-motion alone (\figref{fig:illustration}). The previous work either introduces additional models to estimate optical flow \cite{Yin2018CVPR, Zou2018ECCV, Chen2019ICCV} and also motion masks \cite{Ranjan2019CVPR, Luo2019PAMI} or assumes a given segmentation mask marking semantic regions \cite{Chen2019ICCV, Guizilini2020ICLR} or instances of objects \cite{Casser2019AAAI} on the image. While separate modelling of residual motion with flow or object motion improves the performance, it also increases the computational cost. In case of given segmentation masks, static objects are also segmented and processed separately. Furthermore, errors in the segmentation can propagate to the scene structure. In this paper, we propose to efficiently learn a decomposition of the scene into components with their own motion. As shown in \figref{fig:illustration}, components correspond to moving regions by successfully separating moving objects.

Inspired by the success of SE3-Nets on point clouds \cite{Byravan2017ICRA, Byravan2018ICRA}, we introduce \textbf{MonoDepthSeg} to segment a scene into a set of components and predict the motion of each component from monocular image sequences. We fix the maximum number of components for a model and assume rigid body motion for each component. Instead of one transformation corresponding to ego-motion, we estimate a set of \textbf{SE}(3) transformations to account for the motion of independently moving objects in the scene as well. For this purpose, we modify the pose network's decoder to estimate a fixed number of poses, \ie one per component. In parallel, a mask decoder learns to segment the image into regions whose motion is encoded by the corresponding pose.
In order to model the close relationship between the structure and the semantics, mask and pose estimations are tightly correlated through a shared encoder. This also provides efficiency in comparison to a separate segmentation network entirely. We apply a regularization on the masks by encouraging a depth ordering between them.
\begin{figure*}[t]
    \centering
    \includegraphics[width=\textwidth]{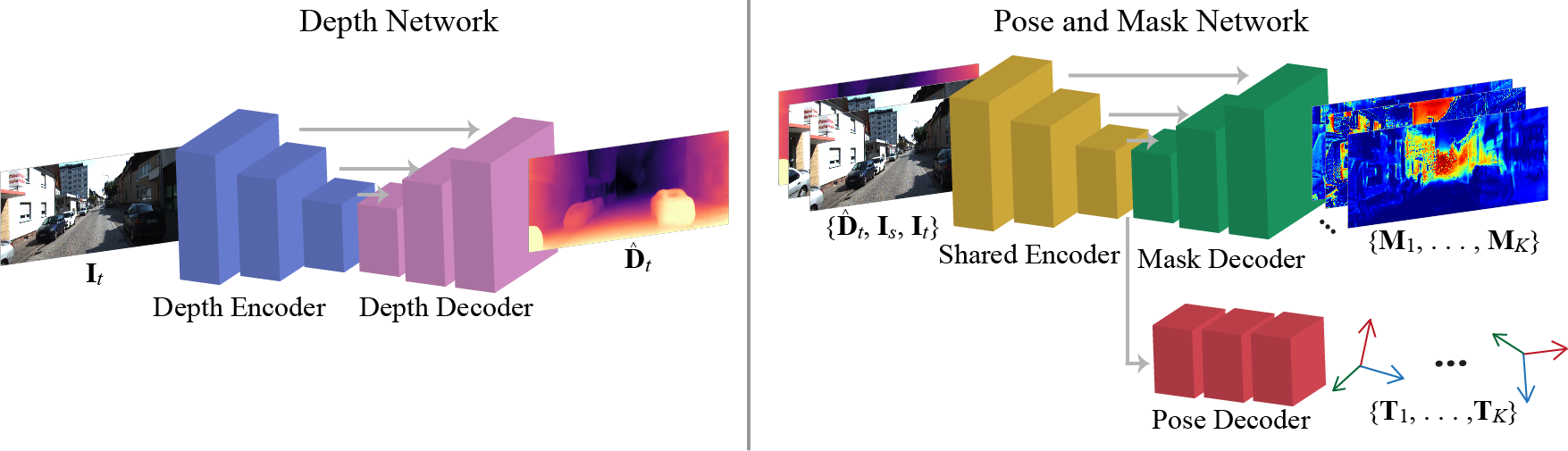}
    \caption{\textbf{MonoDepthSeg.}
    \textit{Left:} The depth network outputs depth estimate $\hat{\bD}_t$ for input $\bI_t$. \textit{Right:} Given $\hat{\bD}_t$ and two consecutive frames $\bI_s$ and $\bI_t$, the shared encoder maps the input frames and the depth estimate to a common representation shared by the following two decoders. The mask decoder produces the same resolution $K$ masks $\{\bM_1,\dots,\bM_K\}$ with skip connections between the corresponding layers of the encoder. The pose decoder takes the same encoded representation and converts it into rigid transformations $\{\bT_1,\dots,\bT_K\}$ corresponding to the masks. Given the set of rigid transformations and corresponding masks, we transform a 3D point as a convex combination of estimated transformations weighted according to estimated masks. The number of components, $K$, is a hyper-parameter of our model.}
    \label{fig:overview}
\end{figure*}

We build on top of the recent self-supervised learning approaches for monocular depth estimation and show the effectiveness of learning to decompose the scene on KITTI~\cite{Geiger2012CVPR, Geiger2013IJRR}, Cityscapes~\cite{Cordts2016CVPR}, and DDAD~\cite{Guizilini2020CVPR} datasets.
Without using an additional network other than the mask decoder, we improve depth estimation and obtain a decomposition of the scene into moving regions which is coherent both semantically and structurally. 
Our extensive experiments demonstrate that our joint formulation is able to resolve problems of the baseline Monodepth2~\cite{Godard2019ICCV}, especially in challenging foreground regions with moving objects on KITTI and significantly outperforms it on more complex Cityscapes and DDAD due to a large number of independently moving objects. In addition, our method is able to extract a decomposition of the scene into moving components which is consistent with the estimated structure~(\figref{fig:illustration}). 

In summary, our contributions are as follows: (1) We introduce a framework for jointly estimating depth and a scene decomposition from monocular video without any ground-truth labels. (2) We estimate the motion of each independently moving object in addition to ego-motion to achieve a more accurate depth estimation.

\section{Related Work}
View synthesis was first proposed by enforcing consistency between two different views in the stereo setup~\cite{Garg2016ECCV}. By using Spatial Transformer Networks (STNs)~\cite{Jaderberg2015NeurIPS}, Monodepth~\cite{Godard2017CVPR} proposes a fully differentiable image reconstruction process by also optimizing for left-right consistency check. According to the depth estimation from a reference view, the source view can be geometrically warped in a differentiable way using STNs to synthesize the reference view. Then, the difference between the synthesized reference view and the original one can be used as a supervisory signal. 
Starting with SfMLearner by Zhou \etal \cite{Zhou2017CVPR}, monocular depth estimation approaches generalize view synthesis to adjacent frames by also estimating the relative pose from one frame to the next. We generalize \cite{Zhou2017CVPR} to scenes with moving objects by applying the idea of $K$ motion models to real-world image sequences that was initially proposed for point clouds in \cite{Byravan2017ICRA}.

\subsection{Self-Supervised Monocular Depth} 
There is an inherent scale ambiguity in monocular depth estimation and several approaches address this issue differently.
Zhan \etal \cite{Zhan2018CVPR} use stereo pairs for joint training of single view depth and two-frame visual odometry. 
Wang \etal \cite{Wang2018CVPR} argue the necessity of handling the scale problem for monocular input and apply an effective normalization on the depth estimation before computing the loss. 
Bian \etal \cite{Bian2019NeurIPS} enforce scale-consistent predictions with a loss minimizing normalized differences in predictions over the entire sequence.
Mahjourian \etal \cite{Mahjourian2018CVPR} propose to go beyond pixel-wise losses by minimizing the alignment error between the inferred 3D geometries of the scene.
There are a number of approaches that propose to jointly estimate monocular depth, optical flow and camera motion. The idea behind these approaches is to handle the camera motion as rigid flow and the remaining non-rigid object motion as residual flow as proposed in GeoNet \cite{Yin2018CVPR}. DF-Net \cite{Zou2018ECCV} proposes to enforce the consistency between depth and flow estimations geometrically in order to eliminate the propagation of errors from pose and depth estimation to flow.
GLNet \cite{Chen2019ICCV} proposes to capture multiple geometric constraints and further improves the results, especially with the epipolar constraint for flow.

Ranjan \etal \cite{Ranjan2019CVPR} learn to segment static parts of the scene where optical flow is estimated using depth and camera motion.
EPC++ \cite{Luo2019PAMI} proposes a holistic 3D motion parser to segment moving objects and estimate motion maps for each as well as the background.
These methods explicitly compute optical flow and motion masks with separate networks at the expense of increased computational complexity. We assume that the most common objects in outdoor scenes such as cars move rigidly but differently than the ego-motion. Based on this, we segment them and model their motion separately, but still in a unified architecture.

Recent approaches \cite{Godard2019ICCV, Guizilini2020CVPR} report improved results without using flow or motion masks. The improvements are due to architectural changes, better loss functions, and some careful consideration of the main assumption in self-supervised monocular training, \ie a moving camera in a static scene. An auto-masking strategy is used to mask out pixels with the same motion as camera, which violate this assumption. Monodepth2 \cite{Godard2019ICCV} proposes to handle occlusions by computing the photometric loss as the minimum over source views instead of averaging which leads to blurred results. In addition, while using multi-scale predictions, photometric error is always computed in the full input resolution to avoid ambiguities at low resolution in low-texture regions.
PackNet \cite{Guizilini2020CVPR} uses 3D convolutions to learn detail-preserving representations via symmetric packing and unpacking blocks.
In this paper, we follow the improvements proposed by recent approaches, specifically Monodepth2 \cite{Godard2019ICCV}, in terms of architecture and loss design. 

\subsection{Depth with Semantics}
Joint training of semantics and depth has been explored before, especially in the context of indoor data where ground-truth depth is available for training \cite{Jiao2018ECCV}. For outdoor scenes, Cao \etal \cite{Cao2019CVPR} propose to predict object motion and depth from stereo motion sequences by using object proposals as input. The proposed method can factor the scene into independently moving objects but require stereo sequences and limited supervision for object bounding boxes. In this paper, our goal is to generalize their success to monocular setup without any supervision. 

For monocular data, Casser \etal \cite{Casser2019AAAI} propose to estimate the motion of each object in addition to the ego-motion based on an initial instance segmentation. Motion estimation and the following warping are performed in masked regions according to given segmentation. We follow a similar approach by estimating the motion of each component but we also estimate the masks instead of relying on an initial segmentation. This way, we save computation not only by removing the need to initially segment the instances but also by focusing on the dynamic instances which require a separate modelling.
While they train separate networks for ego-motion and object motion, our motion estimation is shared across components, leading to better efficiency.

Another two-stage training process for incorporating semantics is proposed by Guizilini \etal \cite{Guizilini2020ICLR}. A pre-trained semantic segmentation network is used to guide the depth network via pixel-adaptive convolutions between the decoders. Semantic-aware features produce better estimations especially for dynamic objects. This proves the tight relationship between the semantics and the depth as we also explore in this paper. While \cite{Guizilini2020ICLR} uses a pre-trained and fixed segmentation network for semantic guidance, we train a pose and mask network to predict corresponding poses and masks, with a shared encoder but separate decoders. Our approach 
can produce masks corresponding to separately moving regions on the image. We also perform a detailed evaluation according to motion segmentation on KITTI and Cityscapes and semantic classes on DDAD and show meaningful improvements for moving regions.


%
%

\section{Methodology}
Our monocular depth and segmentation network, \textbf{MonoDepthSeg}, 
jointly estimates the structure and a decomposition of the scene into moving components from monocular video sequences without using any labels. An overview of our approach is presented in \figref{fig:overview}. Our solution couples these two related tasks efficiently within a single framework.
Given a source frame $\bI_s$ and a target frame $\bI_t$ as input where $\bI_s, \bI_t \in \nR^{W \times H \times 3}$, we decompose the scene into $K$ components where each component is represented with a mask and a rigid body motion:
\begin{equation}
    \hat{\bI}_s = \theta(\bI_s, \bI_t)~~\vert~~\theta \coloneqq \{\bR_i, \bt_i, \bM_i\},~i = 1 \dots K
\end{equation}
where $ \hat{\bI}_s$ is the warped source image according to estimated parameters $\theta$. For $K$ components, $\theta$ denotes the set of rigid body transformations $[\bR_i, \bt_i] \in \textbf{SE}(3)$ and corresponding masks $\bM_i \in [0,1]^{W \times H}$ representing the component $i$. In our setting, we assume that $K$ is a hyper-parameter which limits the number of components that can be estimated.

Our framework consists of two parts: \begin{enumerate*}[label=(\roman*)] \item a depth network which estimates per-pixel depth values \item a pose and mask network \end{enumerate*} which divides the image into components and estimates a separate pose for each component. Finally, every pixel on the target frame is transformed according to estimated parameters, \ie inverse warped by sampling from the source frame, and compared to the original target frame.
The two networks are trained jointly and end-to-end.

\subsection{Scene Decomposition}
\label{sec:scene_decomp}
The depth network estimates the depth $\hat{\bD}_t$ for every pixel on the target image $\bI_t$. The pose network solves two related problems based on a shared representation, \ie identifying components that move together and estimating transformation parameters for each component.
We use the depth network of Monodepth2 \cite{Godard2019ICCV} as explained in \secref{sec:arch}. In this section, we focus on the pose and mask network starting with the assumptions in our model.

In this work, we make two assumptions. First, we assume rigidly moving components whose motion can be explained with a rigid body transformation. This assumption, as explored before \cite{Casser2019AAAI}, typically holds in outdoor scenes where most of the moving objects are rigidly moving vehicles. Otherwise, it is an approximation for deformable objects such as pedestrians.
Second, we have a pre-defined maximum number of components, $K$. Based on the second assumption, we can estimate the segmentation mask $\bM$ by assigning each pixel into one of the $K$ components. In order to make this process differentiable, we assign soft weights by allowing pixels to belong to more than one component.

Formally, $\bM_i(\bp)$ denotes the probability of the pixel $\bp$ belonging to the component $i$ where the sum of all possible $K$ values for the pixel $\bp$ is equal to 1:
\begin{equation}
    \bM(\bp) = \{\bM_1(\bp), \dots, \bM_K(\bp) \}~\vert~ \sum_{i=1}^K \bM_i(\bp) = 1
\end{equation}

Similar to \cite{Gadelha2019ICCV, Sabour2021ARXIV}, we encourage masks to be layered according to a pre-defined depth order, \eg the first mask appears first, then the second, and so on. The ordering helps to account for occlusions by putting the foreground objects before the background. We assign a scalar number $d_i$ to each mask from $1$ to $K$ denoting its order. Then, we weight the mask logits by this number before applying softmax across $K$ channels at every pixel:
\begin{equation}
    \bM_i(\bp) = \frac{\mathrm{e}^{d_i \bM^{'}_i(\bp)}}{\sum_{j=1}^{K} \mathrm{e}^{d_j \bM^{'}_j(\bp)}}
\end{equation}
where $\bM^{'}$ is the output of the mask decoder. With the help of this regularization, probabilities over $K$ values are ordered according to a pre-defined depth ordering.
As a baseline to the depth ordering, we experiment with another regularization to reduce the noise in the masks. We simply encourage mask values to be locally smooth by using an edge-aware smoothness term in the loss function.

%
%

%
%
%

\subsection{Pose Estimation}
We represent the motion of each component using a 3D rigid transformation $\bT=[\bR, \bt] \in \textbf{SE(3)}$ which is composed of a rotation matrix $\bR \in \textbf{SO(3)}$ and a translation vector $\bt \in \nR^3$.
We first shortly introduce the preliminaries regarding the back-projection of a pixel into 3D, the application of transformation $\bT$ in 3D, and the projection of the transformed 3D point into the other view. After introducing the general framework for establishing correspondences between different viewpoints under a rigid transformation, we generalize it to our framework with a set of rigid transformations and corresponding masks. 

\boldparagraph{Preliminaries} Assuming a known intrinsic camera matrix $\bK$, for a pixel $\bp$ on the target image $\bI_t$ and its depth value on $\hat{\bD}_t$, the corresponding 3D point $\bx$ can be computed as follows:
\begin{equation}
    \label{eq:backproject}
    \bx = 
             \hat{\bD}_t(\bp)~\bK^{-1}~\bp
\end{equation}
Given an estimated rigid transformation $\bT = [\bR, \bt]$, we can transform the 3D point $\bx$ and project it to find the corresponding point $\bp'$ on the source image $\bI_s$ as follows:
\begin{align}
    \label{eq:tform}
    \bx' = \bT ~ \bx = \bR~\bx + \bt && \hfill
    \bp' =  \bK   ~ \bx'
\end{align}
All operations are performed in homogeneous coordinates which are removed for simplicity.

\boldparagraph{Decomposition of Transformation}
In this paper, we estimate $K$ rigid transformations corresponding to the motion of $K$ components. We build a tight relationship between the region of a component on the image, \ie the mask, and the motion of that component through a shared encoder. Based on this shared representation, there are two separate decoders for predicting the masks and the transformations. 

Given the set of predicted rigid transformations $\{\bT_i\}$ and corresponding masks $\{\bM_i\}$, we write \eqref{eq:tform} as a weighted combination of $K$ predictions in a differentiable manner:
\begin{equation}
    \label{eq:decomp_tform}
    \bx' = \sum_{i=1}^K \bM_i(\bp)~\bT_i ~ \bx = \sum_{i=1}^K \bM_i(\bp)~(\bR_i ~ \bx + \bt_i)
\end{equation}
where $\bx'$ is the transformed 3D point $\bx$ corresponding to the pixel $\bp$ on the target image. As stated in \cite{Byravan2017ICRA}, the resulting transformation as a convex combination of the transformed 3D points is not in \textbf{SE(3)} anymore. However, rigid transformations as frequently observed in outdoor scenes can be approximated using \eqref{eq:decomp_tform} as shown in our experiments.

\subsection{Self-Supervised Training}
For a pixel $\bp$ on the target image $\bI_t$, we first obtain the corresponding 3D point $\bx$ as shown in \eqref{eq:backproject}. Then, we transform the 3D point $\bx$ to $\bx'$ according to the set of transformations and masks predicted by the decoders as shown in \eqref{eq:decomp_tform}. Finally, we project the transformed 3D point $\bx'$ to the corresponding pixel $\bp'$ on the source image. We repeat this process for all the pixels on the target image.

Following the monocular depth estimation approaches \cite{Zhou2017CVPR}, we use a differentiable inverse warping process to reconstruct the target image $\bI_t$ by sampling pixels from the source image $\bI_s$. Since the projected pixel coordinates are continuous, we perform bilinear interpolation by using differentiable bilinear sampling mechanism proposed in the Spatial Transformer Networks \cite{Jaderberg2015NeurIPS}. This way, we can approximate the value of the warped image $\hat{\bI}_s$ at location $\bp$ in a differentiable manner. As a result, we obtain the corresponding points on the target image $\bI_t$, \ie the warped source image $\hat{\bI}_s$, and the source image $\bI_s$:
\begin{equation}
    \bI_t(\bp) \approx \hat{\bI}_s(\bp) = \bI_s(\bp')
\end{equation}

Our loss function minimizes the difference between the corresponding points over all pixels for pairs of images from a target view and a couple of source views, \ie the previous and the next frame.
For the photometric reconstruction loss, we follow \cite{Godard2019ICCV} and use per-pixel minimum reprojection loss, and also incorporate the structural similarity (SSIM) loss \cite{Wang2004TIP} by setting $\alpha$ to $0.85$:
\begin{equation}
\begin{split}
    \cL_{\textrm{photo}}(\bp) = \min_s \Bigl[ & \left( 1-\alpha\right)\lvert \bI_t(\bp)-\hat{\bI}_s(\bp)\rvert  \\  & + \frac{\alpha}{2}\left(1-\textrm{SSIM} (\bI_t, \hat{\bI}_s )(\bp) \right)  \Bigr]
\end{split}
\end{equation}
$\cL_{\textrm{photo}}$ is minimized over the two source views considered, \ie the previous and the next frames.

Similar to \cite{Wang2018CVPR, Godard2019ICCV}, we define an edge-aware smoothness loss $\cL_{smooth}$ over the mean-normalized inverse depth values.
Our final loss $\cL$ is a weighted sum of the loss functions defined over all pixels, divided by $N$, the number of pixels:
\begin{equation}
   \cL = \frac{1}{N} \sum_{\bp} \cL_{photo}(\bp) + \lambda~ \cL_{smooth}(\bp) 
\end{equation}

\subsection{Network Architecture}
\label{sec:arch}
Depth network is based on the U-Net architecture \cite{Ronneberger2015MICCAI}, where we have an encoder-decoder network with skip connections. The encoder for the depth network is a ResNet50 architecture \cite{He2016CVPR}, which is pre-trained on ImageNet \cite{Russakovsky2015IJCV}. We use the same depth decoder proposed in Monodepth2 \cite{Godard2019ICCV} where the output layer is a sigmoid converted to depth.

Given two consecutive frames, $\bI_s$ and $\bI_t$, and the depth estimate $\hat{\bD}_t$ for $\bI_t$, a second ResNet50 architecture is modified to accept two RGB images and the depth estimate as input. We concatenate the depth estimate as input to the pose and mask network similar to \cite{Li2020ARXIV}.
As shown on the right in \figref{fig:overview}, the output of this encoder is shared between the pose and mask decoders \cite{Byravan2017ICRA}. This way, we learn a shared representation for the structure and the decomposition of the scene in an efficient way.
The pose decoder outputs $K$ 6-DoF relative poses corresponding to $K$ masks. Following \cite{Wang2018CVPR, Godard2019ICCV}, we predict the rotation in axis-angle representation, and scale the rotation and translation outputs by $0.01$.

Based on the same encoded representation as the pose decoder, the mask decoder outputs $K$ masks, \ie a $K$ channel image where each channel represents the probability of each pixel belonging to that component \cite{Byravan2017ICRA}. We use transposed convolutions in the mask decoder to compute pixel-wise masks at the input resolution. We also add skip connections between the encoder and the mask decoder to better preserve boundaries.
For the pose and mask network, we also experiment by initializing it from a pre-trained segmentation model with ResNet50 backbone, DeepLabv3+ \cite{Chen2018ECCV}.

\section{Experiments}
\begin{table*}[t]
    \centering
    \begin{tabular}{c | c c c c | c c c }
\textbf{Components}  & \multicolumn{4}{c}{Lower Better} & \multicolumn{3}{c}{Higher Better} \\
($K$)  & Abs Rel & Sq Rel & RMSE & RMSE$_{log}$ & $\delta < 1.25$ & $\delta < 1.25^2$ & $\delta < 1.25^3$ \\ \hline
2 & 0.091 & 0.832 & 4.092 & 0.167 & 0.919 & \textbf{0.967} &  \underline{0.982}  \\
3 & 0.090 & 0.801 & 4.123 & 0.167 & \underline{0.920} & \textbf{0.967} &  \underline{0.982}  \\
5 & \textbf{0.084} & \underline{0.719} & \textbf{3.943} & \textbf{0.163} & \textbf{0.925} & \textbf{0.967} &  \textbf{0.983}  \\
10 & \underline{0.085} & \textbf{0.620} & \underline{3.984} & \underline{0.164} & 0.917 & \underline{0.966} &  \textbf{0.983}  \\
\end{tabular}

    \vspace{2pt}
    \caption{\textbf{Number of Components.} This table shows our results according to changing the number of components ($K$). The best in each column is shown in bold and the second best is underlined. Increasing the number of components to $K = 5$ performs favorably compared to smaller number of components with $K = 2$ and $K = 3$ and a large number of components with $K = 10$.}
    \label{tab:ablation_K}
\end{table*}
\begin{table*}[t]
    \centering
    \begin{tabular}{c | c | c c c c | c c c }
\textbf{Components} & \textbf{Depth Ordering} & \multicolumn{4}{c}{Lower Better} & \multicolumn{3}{c}{Higher Better} \\
($K$) &  & Abs Rel & Sq Rel & RMSE & RMSE$_{log}$ & $\delta < 1.25$ & $\delta < 1.25^2$ & $\delta < 1.25^3$ \\ \hline
5 & No & 0.089 & 0.832 & \underline{3.983} & \underline{0.164} & \underline{0.924} & \textbf{0.967} &  \underline{0.982}  \\
10 & No & 0.089 & 0.827 & 4.044 & 0.166 & 0.922 & \underline{0.966} &  \underline{0.982}  \\
5 & Yes & \textbf{0.084} & \underline{0.719} & \textbf{3.943} & \textbf{0.163} & \textbf{0.925} & \textbf{0.967} &  \textbf{0.983}  \\
10 & Yes & \underline{0.085} & \textbf{0.620} & 3.984 & \underline{0.164} & 0.917 & \underline{0.966} &  \textbf{0.983}  \\
\end{tabular}
    \vspace{2pt}
    \caption{\textbf{Depth Ordering.} This table shows the results of using depth ordering compared to simply encouraging smooth masks for regularization. %
    Depth ordering improves the performance of our models over all metrics, especially for squared relative error (Sq Rel).
    }
    \label{tab:ablation_depth_ordering}
\end{table*}
\begin{table*}[t]
    \centering

\begin{tabular}{c | c | c c c c | c c c }
\textbf{Components} & \textbf{DeepLabv3+} & \multicolumn{4}{c}{Lower Better} & \multicolumn{3}{c}{Higher Better} \\
($K$) & \textbf{Pre-Training} \cite{Chen2018ECCV}  & Abs Rel & Sq Rel & RMSE & RMSE$_{log}$ & $\delta < 1.25$ & $\delta < 1.25^2$ & $\delta < 1.25^3$ \\ \hline
5 & No & 0.087 & 0.697 & 4.031 & 0.169 & \underline{0.918} & 0.964 &  \underline{0.981}  \\
10 & No & 0.092 & \textbf{0.595} & 4.111 & 0.174 & 0.908 & 0.961 &  \underline{0.981}  \\
5 & Yes & \textbf{0.084} & 0.719 & \textbf{3.943} & \textbf{0.163} & \textbf{0.925} & \textbf{0.967} &  \textbf{0.983}  \\
10 & Yes & \underline{0.085} & \underline{0.620} & \underline{3.984} & \underline{0.164} & 0.917 & \underline{0.966} &  \textbf{0.983}  \\
\end{tabular}
    \vspace{2pt}
    \caption{\textbf{Pre-trained Segmentation Network.} This table compares the results of using a pre-trained DeepLabv3+ model for semantic segmentation \cite{Chen2018ECCV} as the pose and mask network. We modify it to add the pose decoder in addition to the mask decoder and train within our framework. For different number of components, using the modified DeepLabv3+ improves the results.}
    \label{tab:ablation_deeplab}
    \vspace{-4pt}
\end{table*}

We evaluate \textbf{MonoDepthSeg} on KITTI \cite{Geiger2013IJRR}, Cityscapes \cite{Cordts2016CVPR}, and DDAD \cite{Guizilini2020CVPR} datasets.
In this section, we provide details of training and present our results. 
We first perform some ablation studies to justify our design choices and to show the effect of different values of hyper-parameters in our framework.
We then present our results both quantitatively and qualitatively. 

\subsection{Datasets}
\boldparagraph{KITTI}  We train our model on the Eigen split \cite{Eigen2014NeurIPS} of KITTI dataset \cite{Geiger2012CVPR, Geiger2013IJRR}. We follow the pre-processing proposed in \cite{Zhou2017CVPR} and remove the static scenes which results in 39810 samples for training, 4424 for validation, and 697 for test set. 
We also present results using the improved ground-truth obtained with the technique proposed in \cite{Uhrig2017THREEDV} to compare our results to the latest approaches \cite{Guizilini2020CVPR}.
In the improved ground-truth, Uhrig \etal \cite{Uhrig2017THREEDV} remove the outliers in the LiDAR scans and increase the density of the ground-truth, resulting in high quality 652 depth maps, especially improving the original ground-truth for dynamic objects.
We train only on KITTI dataset, without pre-training on Cityscapes as done by some previous work, and compare to the other approaches under the same training conditions. 

\boldparagraph{Cityscapes} We also train and evaluate on the Cityscapes dataset~\cite{Cordts2016CVPR}. For training, we choose 12 of 18 cities containing 30-frame videos from the training split, which results in 48802 samples. For validation, we choose 2 of the 3 cities containing 30-frame videos from the validation split, resulting in 6058 samples. We evaluate on all 1525 samples in the original test set. We use the provided disparity ground-truths obtained using the SGM stereo method \cite{Hirschmuller2007PAMI}, and convert them to depth for evaluation. Similar to KITTI, we evaluate depth for distances up to 80~m.  

\boldparagraph{DDAD} We also conduct experiments on the recent DDAD dataset~\cite{Guizilini2020CVPR} capturing precise structure across images at longer ranges up to 250~m, making it a more challenging and realistic benchmark for the task of depth estimation. We use 12350 training and 3850 validation samples from the front camera and evaluate on 3080 test images using the online evaluation server~\cite{DDADChallenge}.  

\subsection{Training Details}
We use Adam optimizer \cite{Kingma2014ARXIV} with parameters $\beta_1 = 0.9, \beta_2=0.999 , \epsilon=10^{-8}$, with a learning rate of $10^{-4}$. For KITTI, we fix the input resolution to $640 \times 192$ for efficiency and compare to recent approaches trained on resolutions not higher than that, and for Cityscapes we resize the images to $512 \times 256$. On both of these datasets, we train for 20 epochs where the learning rate is reduced to $10^{-5}$ for the last 5 epochs. For DDAD, we use  images of size $640 \times 384$ (similar to \cite{Guizilini2020CVPR}) and train for 30 epochs. 

The weighting in the loss, \ie $\lambda$ is set to $0.001 \times \text{scale}$ where $\text{scale} \in \{1, \frac{1}{2}, \frac{1}{4}, \frac{1}{8} \}$. We use depth ordering for regularization by default, but also experiment with local smoothing of masks. We use the ResNet50 architecture \cite{He2016CVPR} by default for both the depth network and the pose and mask network. For the pose and mask network, we experiment by using the pre-trained DeepLabv3+ architecture for the mask part \cite{Chen2018ECCV}.
We note that architectural improvements \cite{Guizilini2020CVPR} or improvements due to higher input resolutions as observed in recent work \cite{Godard2019ICCV, Guizilini2020CVPR} are complementary to our approach and can be used to improve the results.

\subsection{Ablation Study}
In this section, we analyze different parts of our model including the maximum number of components, depth ordering for regularization, and using a pre-trained segmentation network for the pose and mask network. We train each model for 20 epochs on KITTI and present the results on the validation set using the original ground truth.

\boldparagraph{Number of Components}
In order to evaluate the effect of scene decomposition, we perform an ablation study by changing the maximum number of components in the model. Specifically, we compare different values of $K$ ranging over $2, 3, 5, \text{and } 10$ components. As can be seen from \tabref{tab:ablation_K}, the number of components affects the results. A small number of components can be the limiting factor for the model as observed for $K = 2$ and $K = 3$. While increasing it to $K = 5$ improves the results for all the metrics, $K = 10$ performs a little worse than $K = 5$. The number of components clearly depends on the complexity of scenes in the dataset. This parameter might need to be adjusted for other datasets with more foreground objects but on KITTI, $K=5$ components results in the best performance in almost all metrics among the values compared. 

\boldparagraph{Depth Ordering}
We investigate the effect of depth ordering on the performance of our model by comparing the results with and without depth ordering. When we remove the depth ordering, we still apply a regularization on the masks with simple mask smoothing in order to reduce the noise. 
As can be seen in \tabref{tab:ablation_depth_ordering}, depth ordering improves the results for both $K=5$ and $K=10$.
The difference between the depth ordering and smoothing masks is local versus global regularization. Depth ordering relates the masks over the channels corresponding to components by forcing similar depth-level pixels to cluster on the same mask while mask smoothing only locally enforces similar values on each mask independently. For this reason, we use depth ordering for regularization in the following experiments.

\boldparagraph{Pre-Trained Segmentation Network}
In order to push the limits for the pose and mask network, we use a successful segmentation model, DeepLabv3+ \cite{Chen2018ECCV}, with the ResNet50 backbone and first pre-train it on Cityscapes dataset \cite{Cordts2016CVPR}. Then, we modify the DeepLabv3+ architecture by adding a pose decoder in addition to the mask decoder. Following our initial design, the encoder is still shared between the pose and the mask decoders. We train our framework with the modified DeepLabv3+ for the pose and mask network. Note that we initialize the encoder and the mask decoder from the DeepLabv3+ model pre-trained in a supervised manner on Cityscapes but our training is still self-supervised.
As shown in \tabref{tab:ablation_deeplab}, using the modified DeepLabv3+ model outperforms the original ResNet50 model for the pose and mask network. In the following experiments, we use the modified DeepLabv3+ model.

\subsection{Results}

\begin{table}[b!]
    \begin{tabular}{l|lcccc}
        \toprule
        \multirow{2}{*}{Data} & {\multirow{2}{*}{Method}} &  \multicolumn{2}{c}{Abs Rel} &  \multicolumn{2}{c}{RMSE}\\
         &  & Moving & All & Moving & All
        \tabularnewline 
        \midrule
        \parbox[t]{7.5mm}{\multirow{2}{*}{{KITTI}}}
        & Md2~\cite{Godard2019ICCV} & 0.143 & \textbf{0.110} & 5.949 & \textbf{4.642}
        \tabularnewline
        & Ours  & \textbf{0.138} & \textbf{0.110} &  \textbf{5.796} & 4.700\tabularnewline
                \midrule
        \parbox[t]{7.5mm}{\multirow{2}{*}{{\vtop{\hbox{\strut City-}\hbox{\strut scapes}}}}}
        & Md2~\cite{Godard2019ICCV} & 0.158 & 0.170 & 8.043 & 8.155
        \tabularnewline
        & Ours  & \textbf{0.143} & \textbf{0.142} & \textbf{7.649} & \textbf{7.361} \tabularnewline
       \bottomrule
    \end{tabular}
    \vspace{2pt}
    \caption{\textbf{Quantitative Results in Moving Regions and Overall.} Comparing our approach to Monodepth2~\cite{Godard2019ICCV} (denoted as Md2) for moving regions and overall on KITTI and Cityscapes datasets.}
    \label{tab:motion}
\end{table}

\begin{table*}[t]
    \centering
    \begin{tabular}{l | l | c | c c c c | c c c }
& {\multirow{2}{*}{\bf Method}} & {\multirow{2}{*}{\bf Supervision}} & \multicolumn{4}{c}{Lower Better} & \multicolumn{3}{c}{Higher Better} \\ 
& & & Abs Rel & Sq Rel & RMSE & RMSE$_{log}$ & $\delta < 1.25$ & $\delta < 1.25^2$ & $\delta < 1.25^3$\\ \hline
\parbox[t]{2mm}{\multirow{6}{*}{\rotatebox[origin=c]{90}{\small{Original \cite{Eigen2014NeurIPS}}}}}
& Godard \etal \cite{Godard2019ICCV} & M & \textbf{0.110} & 0.831 & \underline{4.642} & \textbf{0.187} & \textbf{0.883} & \textbf{0.962} & \textbf{0.982} \\
& Guizilini \etal \cite{Guizilini2020CVPR} & M & \underline{0.111} & \textbf{0.785}   & \textbf{4.601} & \underline{0.189} & 0.878 & \underline{0.960} & \textbf{0.982}\\ 
& MonoDepthSeg (Ours) & M & \textbf{0.110} & \underline{0.792} & 4.700 & \underline{0.189} & \underline{0.881} & \underline{0.960} & \textbf{0.982} \\ \cline{2-10}
& Casser \etal \cite{Casser2019AAAI} & S+Ins & 0.141 & 1.025 & 5.290 & 0.215 & 0.816 & 0.945 & 0.979 \\
& Chen \etal \cite{Chen2019ICCV} & S+Sem & 0.118 & 0.905 & 5.096 & 0.211 & 0.839 & 0.945 & 0.977 \\
& Guizilini \etal \cite{Guizilini2020ICLR} & M+Sem & 0.102 & 0.698 & 4.381 & 0.178 & 0.896 & 0.964 & 0.984 \\ \hline\hline
\parbox[t]{2mm}{\multirow{4}{*}{\rotatebox[origin=c]{90}{\small{Improved \cite{Uhrig2017THREEDV}}}}} 
& Luo \etal \cite{Luo2019PAMI} & M & 0.120 & 0.789 & 4.755 & 0.177 & 0.856 & 0.961 & 0.987 \\
& Godard \etal \cite{Godard2019ICCV} & M & \underline{0.085} & 0.468 & \underline{3.672} & \underline{0.128} & \underline{0.921} & \underline{0.985} & \underline{0.995} \\
& Guizilini \etal \cite{Guizilini2020CVPR} & M & \textbf{0.078} & \textbf{0.420} & \textbf{3.485} & \textbf{0.121} & \textbf{0.931} & \textbf{0.986} & \textbf{0.996} \\
& MonoDepthSeg (Ours) & M & \underline{0.085} & \underline{0.458} & 3.779 & 0.131 & 0.919 & \underline{0.985} & \textbf{0.996} \\
\end{tabular}
    \vspace{2pt}
    \caption{\textbf{Quantitative Results on KITTI.} This table compares our proposed approach \textbf{MonoDepthSeg} to previous approaches on KITTI dataset. 
    The \textbf{Supervision} column shows different types of supervision used by different methods, \textbf{M} stands for monocular sequences, and \textbf{S} for methods that train on stereo pairs. We mark the methods which require semantic (\textbf{Sem}) or instance-level (\textbf{Ins}) information. We show the results for the input resolution $640 \times 192$, with pre-trained segmentation network, and by training on KITTI for depth. The best self-supervised method in each column is shown in bold and the second best is underlined.}
    \label{tab:overview}
    \vspace{-2pt}
\end{table*}

\begin{table}[h]
    \vspace{-3mm}
    \begin{tabular}{l|lccccc}
        \toprule
        Method &  Abs Rel  & RMSE & Car & Person\tabularnewline 
        \midrule
        Guizilini \etal~\cite{Guizilini2020CVPR} & 0.23  & 17.92 & 0.38 & 0.20 \tabularnewline
        Godard \etal~\cite{Godard2019ICCV} & 0.22  & 17.63 & 0.25 & 0.21
        \tabularnewline
        Ours  & \textbf{0.19} & \textbf{16.61} & \textbf{0.24} & \textbf{0.17} \tabularnewline
       \bottomrule
    \end{tabular}
    \vspace{1pt}
        \caption{\textbf{Quantitative Results on DDAD.} The \emph{Car} and \emph{Person} are shown using \emph{Abs Rel}. Our method achieves the state-of-the-art results without using a pre-trained segmentation network.}
        \label{tab:ddad}
\end{table}

\begin{figure*}[t]
    \def\imgw{0.245\linewidth}
    \def\imghddad{0.09\linewidth}
    \def\imghcs{0.08\linewidth}
    \def\hspacing{0.1cm}
    \def\vspacing{0.05cm}
    \input{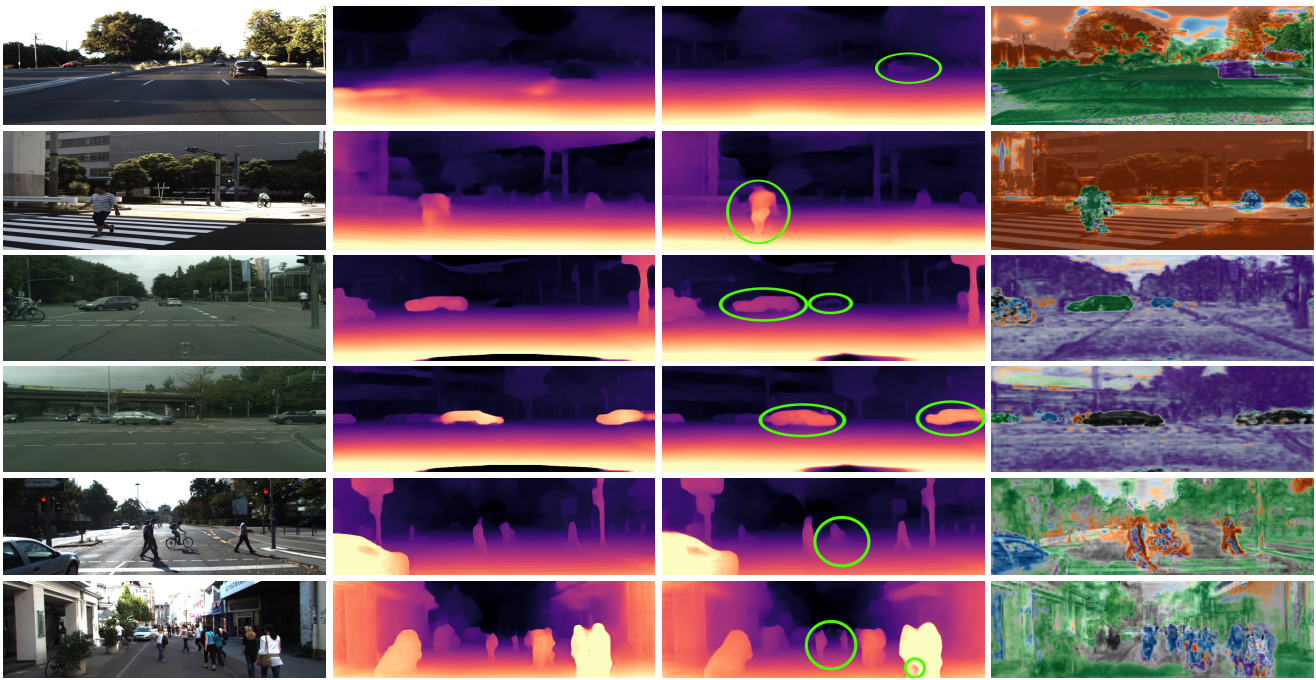}
    \caption{{\bf Qualitative Results.} Each row shows from top-to-bottom: The input image (the first column), Monodepth2 \cite{Godard2019ICCV} results as the baseline (the second column), our results with decomposition (the last two columns).
    In the last column, each channel of the mask is assigned a random color and the brightness level of the color is adjusted according to our estimation. The moving regions (circled in {\color{green}{green}}) containing vehicles, bikes, and pedestrians are clearly segmented out as a separate component and assigned a different color. By learning a separate transformation for each component, we obtain more accurate depth estimations especially in moving foreground regions. 
    }
    \label{fig:qualitative_results}
    \vspace{-3.5pt}
\end{figure*}

We compare the performance of our method to the state-of-the-art by choosing the best performing configuration according to our ablation study, \ie $K = 5$ with depth ordering and pre-trained segmentation network. For KITTI, we evaluate our method both with respect to the original ground-truth~\cite{Eigen2014NeurIPS} and the improved ground-truth~\cite{Uhrig2017THREEDV}. 
As shown in \tabref{tab:overview}, our method performs comparably to the state-of-the-art methods PackNet~\cite{Guizilini2020CVPR} and Monodepth2~\cite{Godard2019ICCV}  in the monocular setting according to both the original and the improved ground-truth.
Furthermore, we generate a scene decomposition jointly without conditioning on any segmentation map.
Our improvements are particularly pronounced in foreground regions with  moving objects (\figref{fig:qualitative_results} and \tabref{tab:motion}) but the overall performance is also improved with respect to the methods which use stereo pairs and rely on the output of a semantic \cite{Chen2019ICCV} or instance segmentation network \cite{Casser2019AAAI}. Furthermore, our joint scene decomposition formulation can compute the masks and poses of images in a single pass (9.057 samples per second), whereas using pre-computed masks as in \cite{Casser2019AAAI}, requires a separate pass for each object in the image (4.124 samples per second). While the state-of-the-art approach~\cite{Guizilini2020CVPR} 
uses an architecture with a large number of parameters due to 3D convolutions,
we achieve similar results with a simpler architecture based on ResNet50. The followup work by the same authors improve the performance further by using supervision for semantic segmentation~\cite{Guizilini2020ICLR}. 
Note that we do not condition on the output of a segmentation network but automatically discover components which move independently~(\figref{fig:qualitative_results}).

We compare our method with Monodepth2 \cite{Godard2019ICCV} in moving regions. We extract moving regions using motion segmentation methods on KITTI~\cite{Mohamed2020ArXiv} and Cityscapes~\cite{Valada2017IROS}.
Our method achieves significantly better results than Monodepth2 on Cityscapes in overall, and clearly outperforms it in moving regions on both KITTI and Cityscapes~(\tabref{tab:motion}). Please check the supplementary for the full tables.

Finally, we evaluate the performance of our model on the more complex DDAD dataset. Note that because the online evaluation server \cite{DDADChallenge} does not allow any semantic supervision, our model is not initialized by a pre-trained segmentation network on this dataset. Despite this, our method outperforms the state-of-the-art method~\cite{Guizilini2020CVPR} and Monodepth2~\cite{Godard2019ICCV} in all metrics including car and person classes, as can be seen in \tabref{tab:ddad}. 

\section{Conclusion and Future Outlook}
We proposed a joint scene decomposition and depth estimation model. We showed that joint formulation improves depth results, especially in moving foreground regions, compared to Monodepth2 as well as the other recent methods which condition on the output of a semantic or instance segmentation. Our model can discover components on the image which correspond to moving regions without any conditioning.
The maximum number of components is fixed, however, outdoor scenes are not equally complex. For most scenes, ego-motion alone can explain the structure of the whole scene, as evidenced by the success of previous monocular depth estimation approaches. As we showed in this paper, their performance can be improved, especially in moving foreground regions, by decomposing the scene and allowing other types of motion estimations. We plan to work on extending our formulation to an adaptive number of components according to scene complexity in the future.\\
{\noindent{\bf Acknowledgements.} S.~Safadoust was supported by KUIS~AI~Center and F.~G\"uney by Marie Skłodowska-Curie Individual and TUBITAK~2232  Fellowship programs.}

\clearpage 
\clearpage 

{\small
\bibliographystyle{ieee_fullname}
\bibliography{bibliography_long, bibliography}

\begin{thebibliography}{10}\itemsep=-1pt

\bibitem{Bian2019NeurIPS}
Jiawang Bian, Zhichao Li, Naiyan Wang, Huangying Zhan, Chunhua Shen, Ming-Ming
  Cheng, and Ian Reid.
\newblock Unsupervised scale-consistent depth and ego-motion learning from
  monocular video.
\newblock In {\em Advances in Neural Information Processing Systems (NeurIPS)},
  pages 35--45, 2019.

\bibitem{Byravan2017ICRA}
Arunkumar Byravan and Dieter Fox.
\newblock {SE3-Nets}: Learning rigid body motion using deep neural networks.
\newblock In {\em Proc. IEEE International Conf. on Robotics and Automation
  (ICRA)}, pages 173--180, 2017.

\bibitem{Byravan2018ICRA}
Arunkumar Byravan, Felix Leeb, Franziska Meier, and Dieter Fox.
\newblock {SE3-Pose-Nets}: Structured deep dynamics models for visuomotor
  control.
\newblock In {\em Proc. IEEE International Conf. on Robotics and Automation
  (ICRA)}, pages 1--8, 2018.

\bibitem{Cao2019CVPR}
Zhe Cao, Abhishek Kar, Christian Hane, and Jitendra Malik.
\newblock Learning independent object motion from unlabelled stereoscopic
  videos.
\newblock In {\em Proc. IEEE Conf. on Computer Vision and Pattern Recognition
  (CVPR)}, pages 5594--5603, 2019.

\bibitem{Casser2019AAAI}
Vincent Casser, Soeren Pirk, Reza Mahjourian, and Anelia Angelova.
\newblock Depth prediction without the sensors: Leveraging structure for
  unsupervised learning from monocular videos.
\newblock In {\em Proc. of the Conf. on Artificial Intelligence (AAAI)}, pages
  8001--8008, 2019.

\bibitem{Chen2018ECCV}
Liang-Chieh Chen, Yukun Zhu, George Papandreou, Florian Schroff, and Hartwig
  Adam.
\newblock Encoder-decoder with atrous separable convolution for semantic image
  segmentation.
\newblock In {\em Proc. of the European Conf. on Computer Vision (ECCV)}, 2018.

\bibitem{Chen2019ICCV}
Yuhua Chen, Cordelia Schmid, and Cristian Sminchisescu.
\newblock Self-supervised learning with geometric constraints in monocular
  video: Connecting flow, depth, and camera.
\newblock In {\em Proc. of the IEEE International Conf. on Computer Vision
  (ICCV)}, pages 7063--7072, 2019.

\bibitem{Cordts2016CVPR}
Marius Cordts, Mohamed Omran, Sebastian Ramos, Timo Rehfeld, Markus Enzweiler,
  Rodrigo Benenson, Uwe Franke, Stefan Roth, and Bernt Schiele.
\newblock The cityscapes dataset for semantic urban scene understanding.
\newblock In {\em Proc. IEEE Conf. on Computer Vision and Pattern Recognition
  (CVPR)}, 2016.

\bibitem{DDADChallenge}
Dense depth for autonomous driving ({DDAD}) challenge.
\newblock 2021.
\newblock https://eval.ai/web/challenges/challenge-page/902/overview.

\bibitem{Eigen2014NeurIPS}
David Eigen, Christian Puhrsch, and Rob Fergus.
\newblock Depth map prediction from a single image using a multi-scale deep
  network.
\newblock In {\em Advances in Neural Information Processing Systems (NeurIPS)},
  pages 2366--2374, 2014.

\bibitem{Gadelha2019ICCV}
Matheus Gadelha, Rui Wang, and Subhransu Maji.
\newblock Shape reconstruction using differentiable projections and deep
  priors.
\newblock In {\em Proc. of the IEEE International Conf. on Computer Vision
  (ICCV)}, 2019.

\bibitem{Garg2016ECCV}
Ravi Garg, Vijay~Kumar Bg, Gustavo Carneiro, and Ian Reid.
\newblock Unsupervised {CNN} for single view depth estimation: Geometry to the
  rescue.
\newblock In {\em Proc. of the European Conf. on Computer Vision (ECCV)}, pages
  740--756, 2016.

\bibitem{Geiger2013IJRR}
Andreas Geiger, Philip Lenz, Christoph Stiller, and Raquel Urtasun.
\newblock Vision meets robotics: The {KITTI} dataset.
\newblock {\em International Journal of Robotics Research (IJRR)}, 2013.

\bibitem{Geiger2012CVPR}
Andreas Geiger, Philip Lenz, and Raquel Urtasun.
\newblock Are we ready for autonomous driving? the kitti vision benchmark
  suite.
\newblock In {\em Proc. IEEE Conf. on Computer Vision and Pattern Recognition
  (CVPR)}, 2012.

\bibitem{Godard2017CVPR}
Cl{\'e}ment Godard, Oisin Mac~Aodha, and Gabriel~J Brostow.
\newblock Unsupervised monocular depth estimation with left-right consistency.
\newblock In {\em Proc. IEEE Conf. on Computer Vision and Pattern Recognition
  (CVPR)}, pages 270--279, 2017.

\bibitem{Godard2019ICCV}
Cl{\'e}ment Godard, Oisin Mac~Aodha, Michael Firman, and Gabriel~J Brostow.
\newblock Digging into self-supervised monocular depth estimation.
\newblock In {\em Proc. of the IEEE International Conf. on Computer Vision
  (ICCV)}, 2019.

\bibitem{Guizilini2020CVPR}
Vitor Guizilini, Rares Ambrus, Sudeep Pillai, Allan Raventos, and Adrien
  Gaidon.
\newblock {3D} packing for self-supervised monocular depth estimation.
\newblock In {\em Proc. IEEE Conf. on Computer Vision and Pattern Recognition
  (CVPR)}, 2020.

\bibitem{Guizilini2020ICLR}
Vitor Guizilini, Rui Hou, Jie Li, Rares Ambrus, and Adrien Gaidon.
\newblock Semantically-guided representation learning for self-supervised
  monocular depth.
\newblock In {\em Proc. of the International Conf. on Learning Representations
  (ICLR)}, April 2020.

\bibitem{Hane2014CVPR}
Christian H{\"{a}}ne, Nikolay Savinov, and Marc Pollefeys.
\newblock Class specific 3d object shape priors using surface normals.
\newblock In {\em Proc. IEEE Conf. on Computer Vision and Pattern Recognition
  (CVPR)}, 2014.

\bibitem{He2016CVPR}
Kaiming He, Xiangyu Zhang, Shaoqing Ren, and Jian Sun.
\newblock Deep residual learning for image recognition.
\newblock In {\em Proc. IEEE Conf. on Computer Vision and Pattern Recognition
  (CVPR)}, pages 770--778, 2016.

\bibitem{Hirschmuller2007PAMI}
Heiko Hirschmuller.
\newblock Stereo processing by semiglobal matching and mutual information.
\newblock {\em IEEE Trans. on Pattern Analysis and Machine Intelligence
  (PAMI)}, 30(2):328--341, 2007.

\bibitem{Hoiem2005ICCV}
D. {Hoiem}, A.~A. {Efros}, and M. {Hebert}.
\newblock Geometric context from a single image.
\newblock In {\em Proc. of the IEEE International Conf. on Computer Vision
  (ICCV)}, 2005.

\bibitem{Jaderberg2015NeurIPS}
Max Jaderberg, Karen Simonyan, Andrew Zisserman, and Koray Kavukcuoglu.
\newblock Spatial transformer networks.
\newblock In {\em Advances in Neural Information Processing Systems (NeurIPS)},
  pages 2017--2025, 2015.

\bibitem{Jiao2018ECCV}
Jianbo Jiao, Ying Cao, Yibing Song, and Rynson Lau.
\newblock Look deeper into depth: Monocular depth estimation with semantic
  booster and attention-driven loss.
\newblock In {\em Proc. of the European Conf. on Computer Vision (ECCV)}, pages
  53--69, 2018.

\bibitem{Kingma2014ARXIV}
Diederik~P Kingma and Jimmy Ba.
\newblock Adam: A method for stochastic optimization.
\newblock {\em arXiv.org}, 1412.6980, 2014.

\bibitem{Ladicky2014CVPR}
Lubor Ladicky, Jianbo Shi, and Marc Pollefeys.
\newblock Pulling things out of perspective.
\newblock In {\em Proc. IEEE Conf. on Computer Vision and Pattern Recognition
  (CVPR)}, 2014.

\bibitem{Li2020ARXIV}
Hanhan Li, Ariel Gordon, Hang Zhao, Vincent Casser, and Anelia Angelova.
\newblock Unsupervised monocular depth learning in dynamic scenes.
\newblock {\em arXiv.org}, 2010.16404, 2020.

\bibitem{Luo2019PAMI}
Chenxu Luo, Zhenheng Yang, Peng Wang, Yang Wang, Wei Xu, Ram Nevatia, and Alan
  Yuille.
\newblock Every pixel counts++: Joint learning of geometry and motion with 3d
  holistic understanding.
\newblock {\em IEEE Trans. on Pattern Analysis and Machine Intelligence
  (PAMI)}, 42(10):2624--2641, 2019.

\bibitem{Mahjourian2018CVPR}
Reza Mahjourian, Martin Wicke, and Anelia Angelova.
\newblock Unsupervised learning of depth and ego-motion from monocular video
  using 3d geometric constraints.
\newblock In {\em Proc. IEEE Conf. on Computer Vision and Pattern Recognition
  (CVPR)}, pages 5667--5675, 2018.

\bibitem{Mohamed2020ArXiv}
Eslam Mohamed, Mahmoud Ewaisha, Mennatullah Siam, Hazem Rashed, Senthil
  Yogamani, and Ahmad El-Sallab.
\newblock Instancemotseg: Real-time instance motion segmentation for autonomous
  driving.
\newblock {\em arXiv preprint arXiv:2008.07008}, 2020.

\bibitem{Ranjan2019CVPR}
Anurag Ranjan, Varun Jampani, Lukas Balles, Kihwan Kim, Deqing Sun, Jonas
  Wulff, and Michael~J Black.
\newblock Competitive collaboration: Joint unsupervised learning of depth,
  camera motion, optical flow and motion segmentation.
\newblock In {\em Proc. IEEE Conf. on Computer Vision and Pattern Recognition
  (CVPR)}, pages 12240--12249, 2019.

\bibitem{Ronneberger2015MICCAI}
Olaf Ronneberger, Philipp Fischer, and Thomas Brox.
\newblock {U-Net}: Convolutional networks for biomedical image segmentation.
\newblock In {\em Medical Image Computing and Computer-Assisted Intervention
  (MICCAI)}, pages 234--241, 2015.

\bibitem{Russakovsky2015IJCV}
Olga Russakovsky, Jia Deng, Hao Su, Jonathan Krause, Sanjeev Satheesh, Sean Ma,
  Zhiheng Huang, Andrej Karpathy, Aditya Khosla, Michael Bernstein, et~al.
\newblock {ImageNet} large scale visual recognition challenge.
\newblock {\em International Journal of Computer Vision (IJCV)},
  115(3):211--252, 2015.

\bibitem{Sabour2021ARXIV}
Sara Sabour, Andrea Tagliasacchi, Soroosh Yazdani, Geoffrey~E. Hinton, and
  David~J. Fleet.
\newblock Unsupervised part representation by flow capsules.
\newblock {\em arXiv.org}, 2011.13920, 2021.

\bibitem{Savinov2016CVPR}
Nikolay Savinov, Christian Haene, Lubor Ladicky, and Marc Pollefeys.
\newblock Semantic {3D} reconstruction with continuous regularization and ray
  potentials using a visibility consistency constraint.
\newblock In {\em Proc. IEEE Conf. on Computer Vision and Pattern Recognition
  (CVPR)}, 2016.

\bibitem{Saxena2007IJCAI}
Ashutosh Saxena, Jamie Schulte, and Andrew Ng.
\newblock Depth estimation using monocular and stereo cues.
\newblock In {\em Proc. of the International Joint Conf. on Artificial
  Intelligence (IJCAI)}, pages 2197--2203, 2007.

\bibitem{Uhrig2017THREEDV}
Jonas Uhrig, Nick Schneider, Lukas Schneider, Uwe Franke, Thomas Brox, and
  Andreas Geiger.
\newblock Sparsity invariant {CNN}s.
\newblock In {\em Proc. of the International Conf. on 3D Vision (3DV)}, 2017.

\bibitem{Valada2017IROS}
Johan Vertens, Abhinav Valada, and Wolfram Burgard.
\newblock Smsnet: Semantic motion segmentation using deep convolutional neural
  networks.
\newblock In {\em Proc. IEEE International Conf. on Intelligent Robots and
  Systems (IROS)}, Vancouver, Canada, 2017.

\bibitem{Wang2018CVPR}
Chaoyang Wang, Jos{\'e} Miguel~Buenaposada, Rui Zhu, and Simon Lucey.
\newblock Learning depth from monocular videos using direct methods.
\newblock In {\em Proc. IEEE Conf. on Computer Vision and Pattern Recognition
  (CVPR)}, pages 2022--2030, 2018.

\bibitem{Wang2004TIP}
Zhou Wang, Alan~C Bovik, Hamid~R Sheikh, and Eero~P Simoncelli.
\newblock Image quality assessment: from error visibility to structural
  similarity.
\newblock {\em IEEE Trans. on Image Processing (TIP)}, 13(4):600--612, 2004.

\bibitem{Yin2018CVPR}
Zhichao Yin and Jianping Shi.
\newblock {GeoNet}: Unsupervised learning of dense depth, optical flow and
  camera pose.
\newblock In {\em Proc. IEEE Conf. on Computer Vision and Pattern Recognition
  (CVPR)}, pages 1983--1992, 2018.

\bibitem{Zhan2018CVPR}
Huangying Zhan, Ravi Garg, Chamara Saroj~Weerasekera, Kejie Li, Harsh Agarwal,
  and Ian Reid.
\newblock Unsupervised learning of monocular depth estimation and visual
  odometry with deep feature reconstruction.
\newblock In {\em Proc. IEEE Conf. on Computer Vision and Pattern Recognition
  (CVPR)}, pages 340--349, 2018.

\bibitem{Zhou2017CVPR}
Tinghui Zhou, Matthew Brown, Noah Snavely, and David~G Lowe.
\newblock Unsupervised learning of depth and ego-motion from video.
\newblock In {\em Proc. IEEE Conf. on Computer Vision and Pattern Recognition
  (CVPR)}, pages 1851--1858, 2017.

\bibitem{Zou2018ECCV}
Yuliang Zou, Zelun Luo, and Jia-Bin Huang.
\newblock {DF-Net}: Unsupervised joint learning of depth and flow using
  cross-task consistency.
\newblock In {\em Proc. of the European Conf. on Computer Vision (ECCV)}, pages
  36--53, 2018.

\end{thebibliography}
}

\clearpage 
\clearpage 

\twocolumn[{%
  \begin{@twocolumnfalse}
    \supptitle{Supplementary Material for \\
Self-Supervised Monocular Scene Decomposition and Depth Estimation}
  \end{@twocolumnfalse}
}]

\def\thesection {\Alph{section}}

\setcounter{section}{0}
\renewcommand{\theHsection}{Supplement.\thesection}
In this Supplementary document, we first provide the details of the architecture in \secref{sec:supp_arch_details}. We then provide additional results to the ones presented in the main paper, both quantitatively (\secref{sec:supp_quantitative}) and qualitatively (\secref{sec:supp_qualitative}). First, we compare the architecture with two separate encoders for pose and mask to our shared encoder. Then, we provide experiments by changing the resolution of input images, with both higher and lower resolutions. At the end of quantitative results, we include full versions of \tabref{tab:motion} and \tabref{tab:ddad} of the main paper which include more results and metrics. Finally in the qualitative results \secref{sec:supp_qualitative}, we show a visual comparison of different number of components and provide more qualitative examples from the best model on all three datasets. Additionally, we provide a video demonstrating the performance of our model on the three datasets. In the video, similar to \figref{fig:illustration} in the main paper, we show raw image (top-left), Monodepth2's depth estimation \cite{Godard2019ICCV} (bottom-left), our depth estimation (bottom-right), and our scene decomposition (top-right). To access the video, please visit \url{https://kuis-ai.github.io/monodepthseg/}
\section{Architecture Details}
\label{sec:supp_arch_details}
\begin{table}[b]
\centering
    \begin{minipage}[b]{\hsize}\centering
    
\begin{tabular}{|c|c|c|c|c|c|c}
\hline
\multicolumn{7}{|l|}{\textbf{Pose Decoder}}                                                                              \\ \hline
\textbf{Layer} & \textbf{k} & \textbf{s} & \textbf{p} & \textbf{Channels} & \textbf{Resolution} & \multicolumn{1}{c|}{\textbf{Act.}} \\ \hline
conv0         & 1          & 1          & 0          & 256           & 1/16         & \multicolumn{1}{c|}{ReLU}         \\ \hline
conv1         & 3          & 1          & 1          & 256           & 1/16         & \multicolumn{1}{c|}{ReLU}         \\ \hline
conv2         & 3          & 1          & 1          & 256           & 1/16         & \multicolumn{1}{c|}{ReLU}         \\ \hline
conv3         & 1          & 1          & 0          & $K \times 6$         & 1/16         & \multicolumn{1}{c|}{-}            \\ \hline
\end{tabular}

    \vspace{8pt}
    \caption{\textbf{Architecture of the Pose Decoder.} The hyper-parameter \textbf{k} is the kernel size, \textbf{s} the stride, \textbf{p} the padding, \textbf{Channels} the number of output channels, \textbf{Resolution} is the image size relative to the input image, and \textbf{Act.} is the activation function used at each layer.}
    \label{tab:supp_architecture}
    \end{minipage}%
\end{table}
In this section, we describe the architecture of our best performing model. 
Our depth network is similar to that of Monodepth2's depth network \cite{Godard2019ICCV} but we use a ResNet50 encoder instead of ResNet18. 
For the shared encoder of the mask and pose network, we modify the Deeplabv3+'s encoder to accept two frames as well as the depth output as input. Therefore, its first layer is a convolutional layer that takes 7 channels as input. The Deeplabv3+ \cite{Chen2018ECCV} is originally pre-trained for semantic segmentation (except for the DDAD \cite{Guizilini2020CVPR} experiment), its last classification layer outputs class probabilities over a fixed number of classes but we need $K$ numbers for our mask prediction. Therefore, we modify the last convolutional layer to output $K$ channels and we initialize its weights randomly. 
In \tabref{tab:supp_architecture}, we describe the architecture of our pose decoder which is similar to the pose decoder used in Monodepth2 except that the number of outputs at the last layer is $K \times 6$ for predicting $K$ 6-DoF poses for each component.

\section{Additional Quantitative Results}
\label{sec:supp_quantitative}
In this section, we provide additional quantitative results on KITTI \cite{Geiger2013IJRR, Geiger2012CVPR} in addition to the ones presented in the main paper by comparing the proposed shared encoder between the pose and mask to separate encoders for each, by changing the image resolution in the paper to both lower and higher resolutions, and finally expanding the Cityscapes \cite{Cordts2016CVPR} and DDAD \cite{Guizilini2020CVPR} results, and the results on the moving regions into more detailed tables.

\boldparagraph{Shared versus Separate Encoders}
%
\begin{table*}[t]
\begin{adjustwidth}{-0.1in}{-0.1in}
    \centering
    \begin{tabular}{c | c | c c c c | c c c }
\textbf{Pose \& Mask} & \multirow{2}{*}{ \textbf{Ground Truth}} & \multicolumn{4}{c}{Lower Better} & \multicolumn{3}{c}{Higher Better} \\
\textbf{Encoder} &  & Abs Rel & Sq Rel & RMSE & RMSE$_{log}$ & $\delta < 1.25$ & $\delta < 1.25^2$ & $\delta < 1.25^3$ \\ \hline
Separate & Original & 0.112 & \textbf{0.747} & 4.855 & 0.194 & 0.870 & 0.957 &  0.981  \\
Shared & Original & \textbf{0.110} & 0.792 & \textbf{4.700} & \textbf{0.189} & \textbf{0.881} & \textbf{0.960} &  \textbf{0.982}  \\ \hline
Separate & Improved & 0.089 & 0.481 & 4.100 & 0.140 & 0.906 & 0.980 &  0.995  \\
Shared & Improved & \textbf{0.085} & \textbf{0.458} & \textbf{3.779} & \textbf{0.131} & \textbf{0.919} & \textbf{0.985} &  \textbf{0.996}  \\
\end{tabular}

    \vspace{8pt}
    \caption{\textbf{Comparison of Shared Encoder and Separate Encoders for Pose and Mask Network on KITTI.} Using separate encoders for pose and mask increases the number of parameters of the model and also reduces the performance according to both the original and improved ground truth on KITTI. This validates our design choice to use a shared encoder for the pose and mask. }
    \label{tab:supp_encoder}
\end{adjustwidth}
\end{table*}
In \tabref{tab:supp_encoder}, we evaluate the effect of separating the encoders for pose and mask in comparison to a shared encoder as we proposed in the main paper. As can be seen from the results, our shared encoder not only reduces the number of parameters, leading to faster training, but also improves the performance by relating the pose and mask predictions. The pose and mask of the components are interleaved with each other through the shared encoder to produce corresponding pose and mask predictions for each component via separate decoders.

\boldparagraph{Effect of Image Resolution}
%
\begin{table*}[t]
\begin{adjustwidth}{-0.1in}{-0.1in}
    \centering

\begin{tabular}{ c | c | c| c c c c | c c c }
{\multirow{2}{*}{\textbf{Resolution}}} & {\multirow{2}{*}{ \textbf{Ground Truth}}} & \textbf{Train Time} &\multicolumn{4}{c|}{Lower Better} & \multicolumn{3}{c}{Higher Better} \\ 
& & \textbf{(hours)} & Abs Rel & Sq Rel & RMSE & RMSE$_{log}$ & $\delta < 1.25$ & $\delta < 1.25^2$ & $\delta < 1.25^3$ \\ \hline
 $416\times128$  & Original &  11 & 0.119 & 0.827 & 5.053 & 0.201 & 0.856 & 0.952 & 0.979 \\
$640\times192$  & Original & 22 &\underline{0.110} & \underline{0.792} & \underline{4.700} & \underline{0.189} & \underline{0.881} & \underline{0.960} & \underline{0.982} \\ 
 $1024\times320$  & Original & 11 + 32 &\textbf{0.107} & \textbf{0.727} & \textbf{4.520} & \textbf{0.184} & \textbf{0.888} & \textbf{0.964} & \textbf{0.983} \\ \hline
$416\times128$  & Improved & 11 &0.094 & 0.553 & 4.328 & 0.149 & 0.896 & 0.974 & 0.993 \\
$640\times192$  &  Improved & 22 &\underline{0.085} & \underline{0.458} & \underline{3.779} & \underline{0.131} & \underline{0.919} & \underline{0.985} & \underline{0.996} \\ 
 $1024\times320$  &  Improved & 11 + 32 & \textbf{0.083} & \textbf{0.405} & \textbf{3.562} & \textbf{0.125} & \textbf{0.925} & \textbf{0.987} & \textbf{0.997} \\ 
\end{tabular}
    \vspace{8pt}
    \caption{\textbf{Effect Of Image Resolution on KITTI.} This table compares the performances and training times for our models using different image resolutions. The timing for the high resolution model comprises 10 epochs of of training of the $640\times 192$ model and 10 epochs of training at $1024 \times 320$ resolution.}
    \label{tab:supp_resolution}
\end{adjustwidth}
\end{table*}
In the main paper, all of our models on KITTI were trained at $640 \times 192$ resolution. In \tabref{tab:supp_resolution}, we show additional results at higher ($1024 \times 320$) and lower ($416 \times 128$) resolutions on this dataset to show the effect of image size on the performance. The low and normal resolution models were trained for 20 epochs. The high resolution model was initialized using the weights from the normal resolution model after 10 epochs of training and was trained for another 10 epochs using a learning rate of $10^{-5}$.
We can see that increasing the resolution improves the performance of the model, but at the expense of longer training time.

\boldparagraph{Full Version of Tables}
In this section, we provide the full versions of \tabref{tab:motion} and \tabref{tab:ddad} of the main paper. \tabref{tab:supp_motion} compares our model to Monodepth2~\cite{Godard2019ICCV} on the moving regions of KITTI and Cityscapes datasets. Our method outperforms Monodepth2~\cite{Godard2019ICCV} on both datasets in all metrics, except for one metric on the KITTI dataset.
In \tabref{tab:supp_cityscapes}, we provide the table with all the metrics to compare our model to Monodepth2~\cite{Godard2019ICCV} on Cityscapes in all regions. Our method performs better than Monodepth2~\cite{Godard2019ICCV} in all metrics. Note that the full results on KITTI in all regions is already provided in \tabref{tab:overview} of the main paper. 

\tabref{tab:supp_ddad} shows the full quantitative results on the DDAD dataset, as reported by the online evaluation server \cite{DDADChallenge} after the challenge ended on June 19, 2021. In this table, we only include the results for the published methods.  Our method achieves state-of-the-art results in all metrics.

\begin{table*}[t]
    \centering
    \begin{tabular}{l|l|cccc|ccc}
        \toprule
        Dataset & Method &  Abs Rel & Sq Rel & RMSE & RMSE$_{log}$ & $\delta < 1.25$ & $\delta < 1.25^2$ & $\delta < 1.25^3$ \tabularnewline 
        \midrule
        \parbox[t]{18mm}{\multirow{2}{*}{{KITTI}}} &
        Godard \etal \cite{Godard2019ICCV} & 0.143 & 1.826 & 5.949 & 0.229 & 0.823 & \textbf{0.913} & 0.951 
        \tabularnewline
        & Ours  & \textbf{0.138} & \textbf{1.709} & \textbf{5.796} & \textbf{0.223} & \textbf{0.837} & 0.912 & \textbf{0.953} \tabularnewline
       \bottomrule
       \parbox[t]{18mm}{\multirow{2}{*}{{Cityscapes}}} &
       Godard \etal \cite{Godard2019ICCV} & 0.158 & 3.148 & 8.043 & 0.241 & 0.820 & 0.940 & 0.972 
        \tabularnewline
        & Ours  & \textbf{0.143} & \textbf{2.018} & \textbf{7.649} & \textbf{0.225} & \textbf{0.824} & \textbf{0.944} & \textbf{0.977} \tabularnewline
       \bottomrule
    \end{tabular}
    %
    \vspace{8pt}
    \caption{\textbf{Full Quantitative Results in Moving Regions on KITTI and Cityscapes.} 
    This table is the expanded version of Table~4 in the main paper for the moving regions, comparing our approach to Monodepth2~\cite{Godard2019ICCV} for moving regions on KITTI and Cityscapes datasets for all metrics. Our method achieves better results than Monodepth2~\cite{Godard2019ICCV} on both datasets in every metric except for one metric on KITTI dataset.}
    \label{tab:supp_motion}
\end{table*} \begin{table*}[t]
    \centering
    \begin{tabular}{l|cccc|ccc}
        \toprule
        Method &  Abs Rel  & Sq Rel & RMSE & RMSE$_{log}$ & $\delta < 1.25$ & $\delta < 1.25^2$ & $\delta < 1.25^3$ \tabularnewline 
        \midrule
         Godard \etal \cite{Godard2019ICCV} & 0.170 & 3.936 & 8.155 & 0.243 & 0.822 & 0.940 & 0.971 
        \tabularnewline
        Ours  & \textbf{0.142} & \textbf{1.942} & \textbf{7.361} & \textbf{0.218} & \textbf{0.827} & \textbf{0.946} & \textbf{0.978} \tabularnewline
       \bottomrule
    \end{tabular}
    \vspace{8pt}
    \caption{\textbf{Full Quantitative Results Overall on Cityscapes.}  This table is the expanded version of Table~4 in the main paper for Cityscapes overall, comparing our approach to Monodepth2~\cite{Godard2019ICCV} on Cityscapes in overall for all metrics. Our method achieves better results than Monodepth2~\cite{Godard2019ICCV} in every metric.}
    \label{tab:supp_cityscapes}
\end{table*}
\begin{table*}[h]
    \centering
    \begin{tabular}{l|ccccccc}
        \toprule
        Method &  Abs Rel  & RMSE & SILog & $\delta < 1.25$ & Car & Person & Semantic Classes (Average)\tabularnewline 
        \midrule
        Guizilini \etal~\cite{Guizilini2020CVPR} & 0.23  & 17.92 & 32.56 & 0.72 &  0.38 & 0.20 & 0.24\tabularnewline
        Godard \etal~\cite{Godard2019ICCV} & 0.22  & 17.63 & 31.93 & 0.70 & 0.25 & 0.21 & 0.24
        \tabularnewline
        Ours  & \textbf{0.19} & \textbf{16.61} & \textbf{29.37} & \textbf{0.75} & \textbf{0.24} & \textbf{0.17} & \textbf{0.22} \tabularnewline
       \bottomrule
    \end{tabular}
    \vspace{8pt}
        \caption{\textbf{Full Quantitative Results on DDAD.} This table is the expanded version of Table~6 in the main paper. The \emph{Car} and \emph{Person} are shown using \emph{Abs Rel}. The last column is the average of \emph{Abs Rel} errors over all of the semantic classes. For every metric, lower is better, except for $\delta < 1.25$, where higher is better.  Our method achieves the state-of-the-art results in every metric, without being initialized by a pre-trained segmentation network.}
        \label{tab:supp_ddad}
\end{table*}

\section{Additional Qualitative Results}
\label{sec:supp_qualitative}
%
%

\begin{figure*}[t]
    \def\imgw{0.195\linewidth}
    \def\hspacing{0.05cm}
    \def\vspacing{0.15cm}
    \input{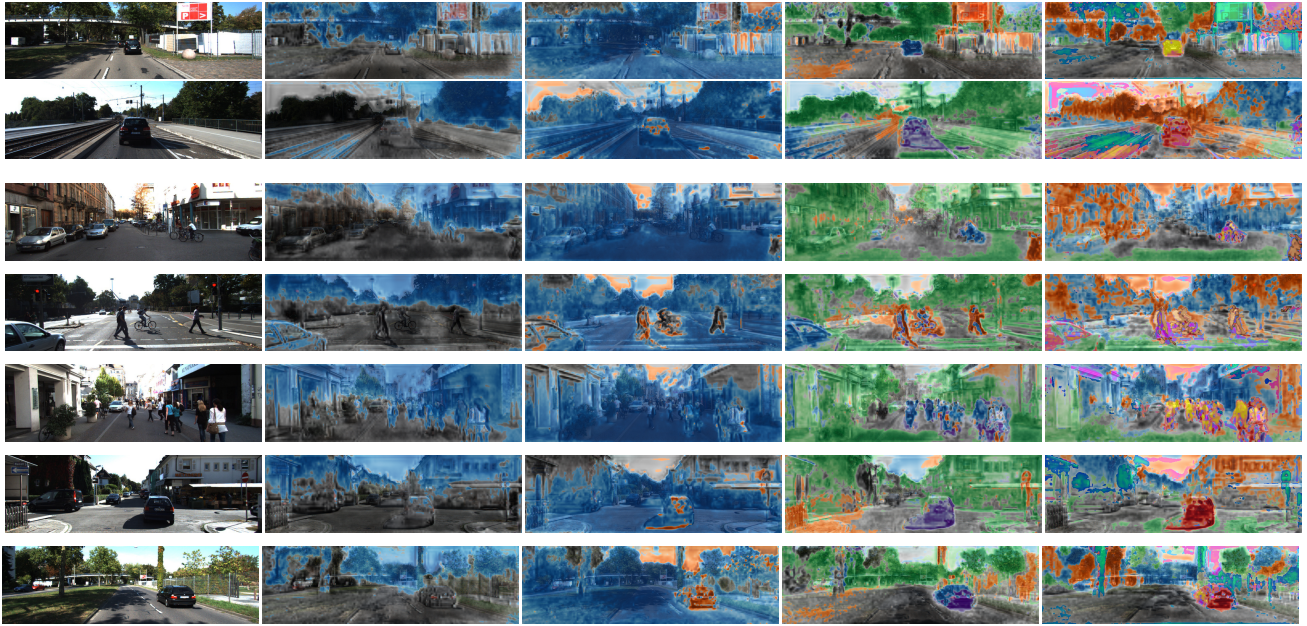}
    \caption{{\bf Effect of $K$ on Masks.} Comparing mask visualizations for different maximum number of components ($K$). Each row shows raw image (first column), and mask visualizations for $K=2, 3, 5, 10$ in the next columns. Small values for $K$ cannot separate the moving objects, while larger values of $K$ can result in a single object being segmented into multiple components.}
    \label{fig:supp_qual_K_masks}
\end{figure*}
\begin{figure*}[t]
    \def\imgw{0.195\linewidth}
    \def\hspacing{0.05cm}
    \def\vspacing{0.15cm}
    \input{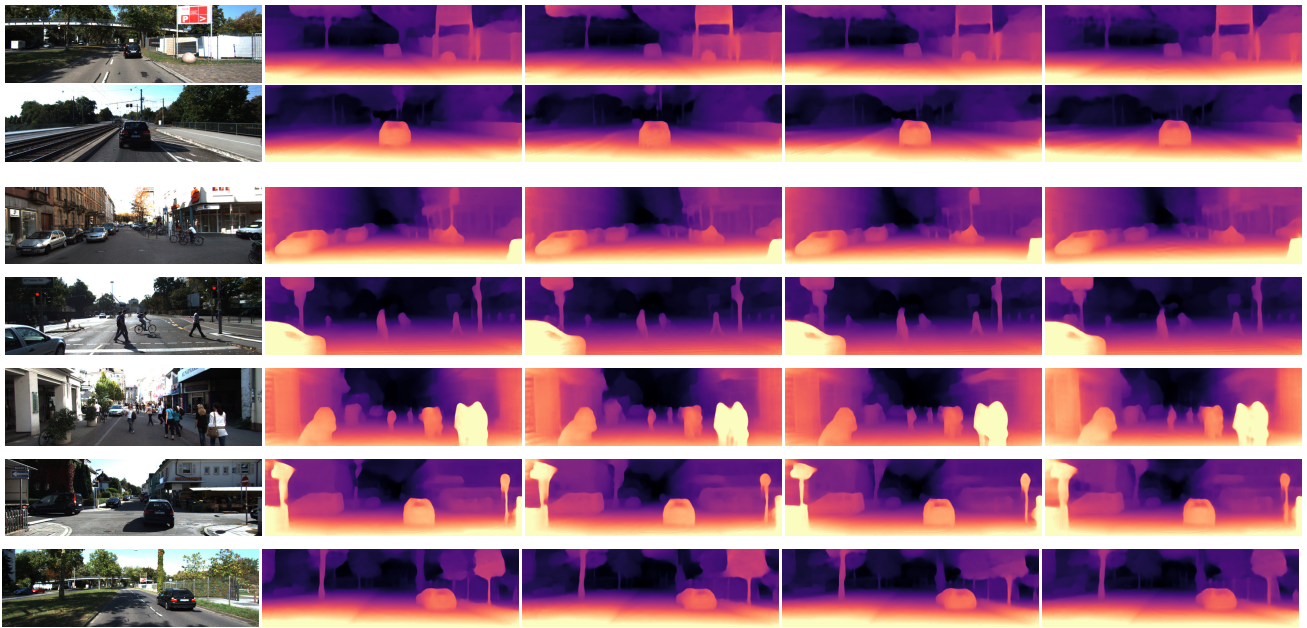}
    \caption{{\bf Effect of $K$ on Depth.} Comparing depth visualizations for different K. Each row shows raw image (first column), and depth visualizations for $K=2, 3, 5, 10$ in the next columns. }
    \label{fig:supp_qual_K_disps}
\end{figure*}

\begin{figure*}[t]
    \def\imgw{0.24\linewidth}
    \def\hspacing{0.1cm}
    \def\vspacing{0.15cm}
    \input{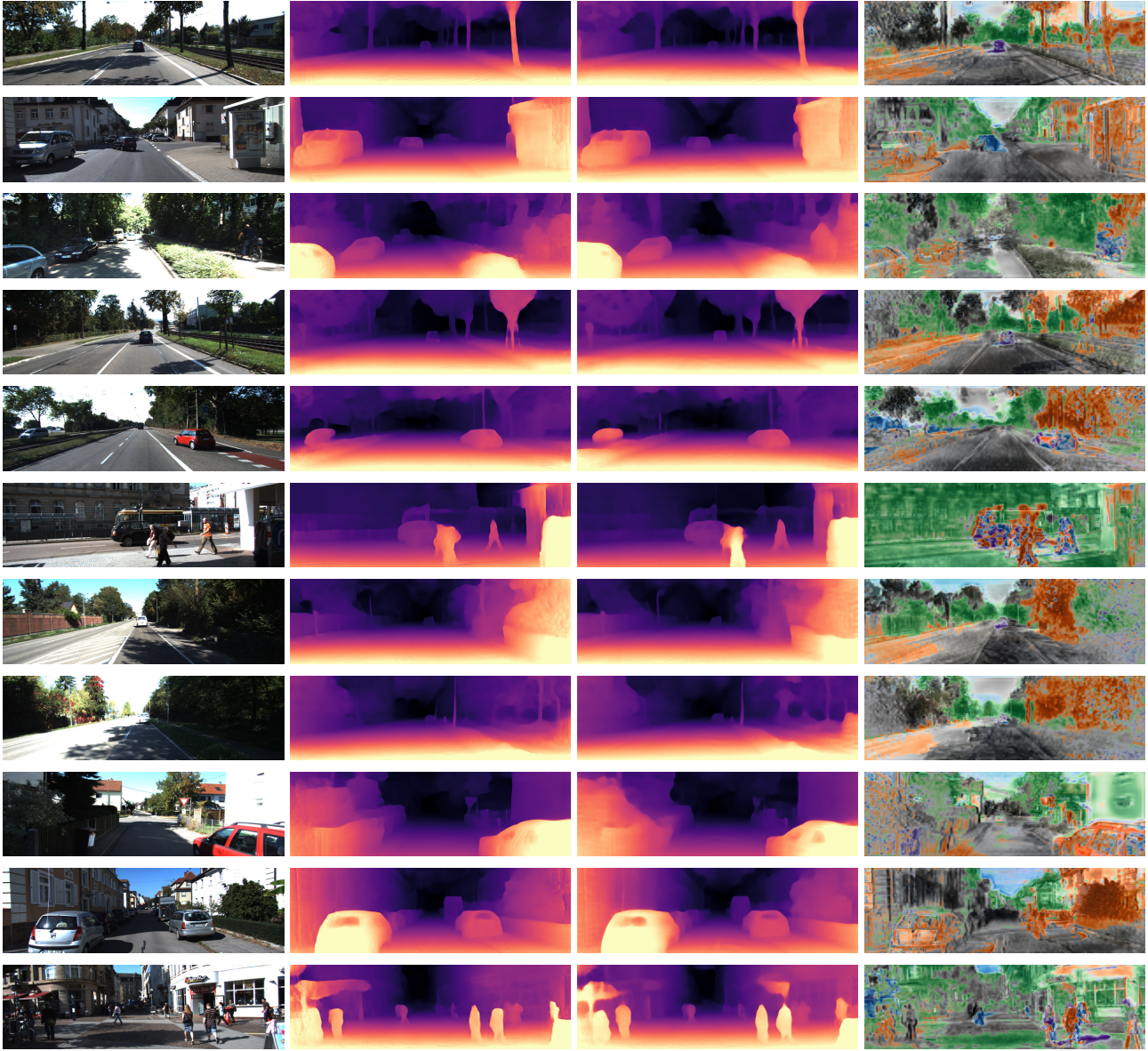}
    \caption{{\bf Additional Qualitative Results on KITTI.} Each row shows from top-to-bottom: The input image (the first column), Monodepth2 results as the baseline (the second column), our results with decomposition (the last two columns). In the last column, each channel of the mask is assigned a random color and the brightness level of the color is adjusted according to our estimation. In the last row it is clear that our model performs better for pedestrians especially for the person at the right.}
    \label{fig:supp_qual_more}
\end{figure*}

\begin{figure*}[t]
    \def\imgw{0.24\linewidth}
    \def\hspacing{0.1cm}
    \def\vspacing{0.15cm}
    \input{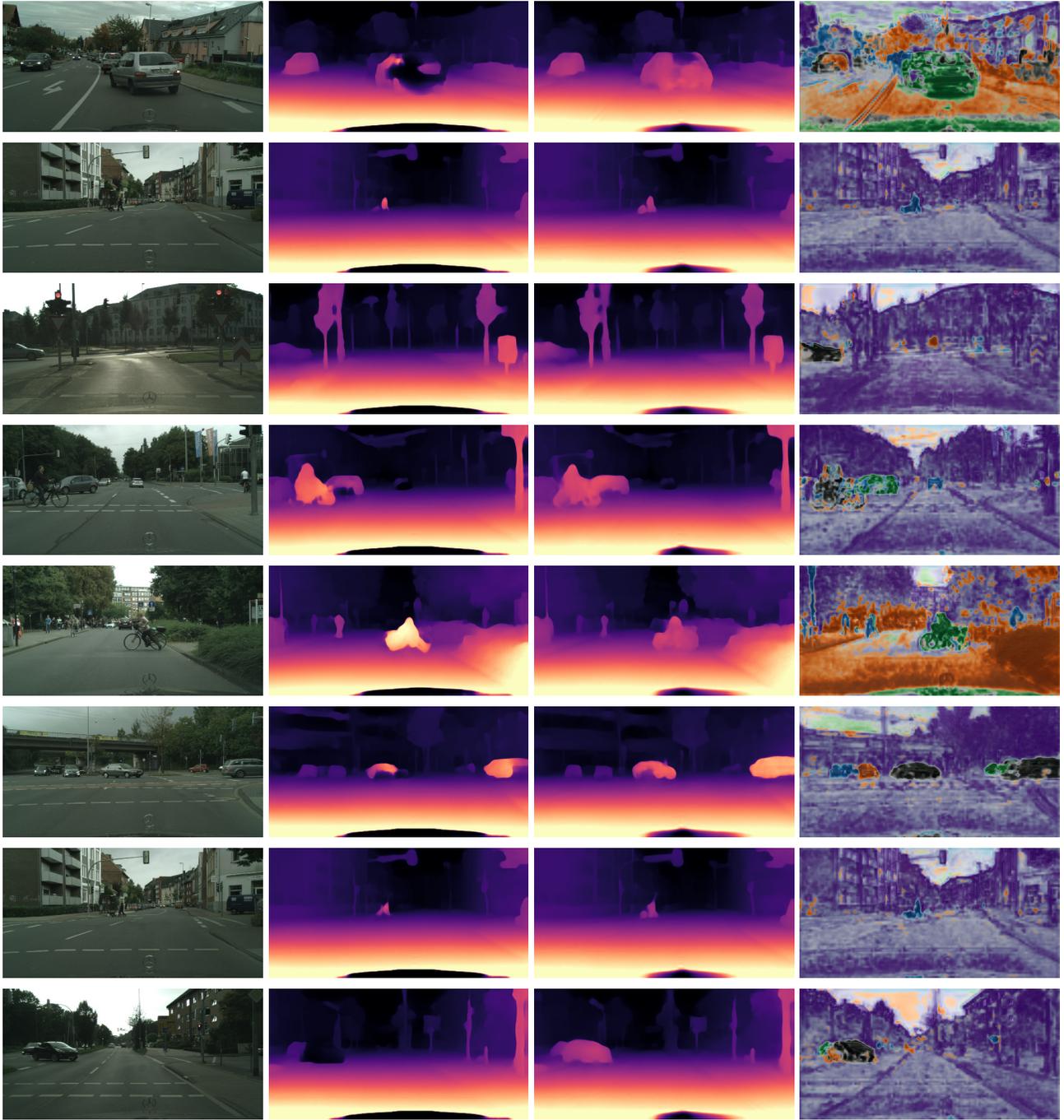}
    \caption{{\bf Additional Qualitative Results on Cityscapes.} Each row shows from top-to-bottom: The input image (the first column), Monodepth2 results as the baseline (the second column), our results with decomposition (the last two columns). In the last column, each channel of the mask is assigned a random color and the brightness level of the color is adjusted according to our estimation.}
    \label{fig:supp_qual_cs}
\end{figure*}

\begin{figure*}[t]
    \def\imgw{0.24\linewidth}
    \def\hspacing{0.1cm}
    \def\vspacing{0.15cm}
    \input{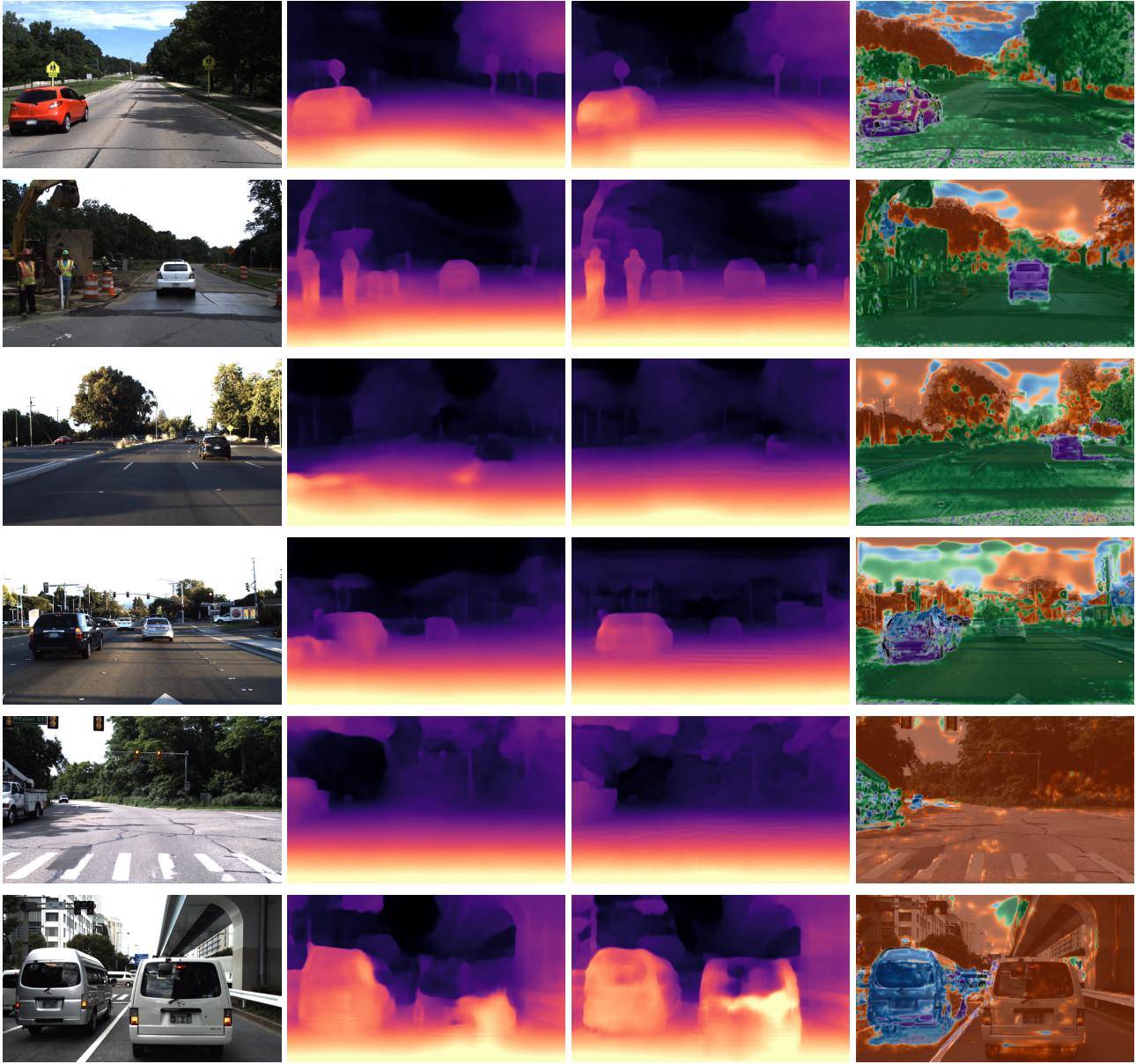}
    \caption{{\bf Additional Qualitative Results on DDAD.} Each row shows from top-to-bottom: The input image (the first column), Monodepth2 results as the baseline (the second column), our results with decomposition (the last two columns). In the last column, each channel of the mask is assigned a random color and the brightness level of the color is adjusted according to our estimation. In the last image, the car on the left is moving while the car on the right is not.}
    \label{fig:supp_qual_ddad}
\end{figure*}

In Figures \ref{fig:supp_qual_more}, \ref{fig:supp_qual_cs}, and \ref{fig:supp_qual_ddad}, we provide additional qualitative results of our best model on KITTI, Cityscapes and DDAD datasets, respectively. We can see that our model performs better especially in moving foreground regions, and at the same time is able to decompose the scene into separately moving components.

\boldparagraph{Number of Components}
To better understand the effect of maximum number of components $K$ on the performance of the model, we include qualitative results comparing the performances of our models using $K=2, 3, 5, 10$. 
In \figref{fig:supp_qual_K_masks}, we compare the scene decomposition results for the same input for different values of $K$. We can see that for $K=2$, the model is unable to separate the moving objects. The scene is divided as the nearby road region and the rest which is further away. With $K = 3$ components, moving objects are better separated but the scene decomposition is not very good because the third component is usually allocated to regions which are impossible to match such as the sky.   
The setting we used in our experiments which is $K=5$ produces the best results by separating the nearby road region, moving objects, and parts of the scene which are further away. As can be seen from the examples, this also provides flexibility when there is more than one moving object.
With $K=10$, a single object might be decomposed into multiple components, leading to slightly worse performance than $K=5$.  
\figref{fig:supp_qual_K_disps} shows the depth estimations for the same set of images for different values of $K$.

\end{document}